\documentclass[letterpaper]{article} 
\usepackage{aaai25}  
\usepackage{times}  
\usepackage{helvet}  
\usepackage{courier}  
\usepackage[hyphens]{url}  
\usepackage{graphicx} 
\usepackage{natbib}  
\usepackage{caption} 
\usepackage{paralist}
\usepackage{algorithm}
\usepackage{algorithmic}
\usepackage{comment}
\usepackage{newfloat}
\usepackage{listings}
\DeclareCaptionStyle{ruled}{labelfont=normalfont,labelsep=colon,strut=off} 
\floatstyle{ruled}
\newfloat{listing}{tb}{lst}{}
\floatname{listing}{Listing}
\usepackage{amsmath}        
\usepackage{amssymb}        
\usepackage{multirow}       
\usepackage{booktabs}       

\title{Enhancing Online Reinforcement Learning with Meta-Learned Objective from Offline Data}
\author{
    Shilong Deng\textsuperscript{\rm 1},
    Zetao Zheng\textsuperscript{\rm 1,2},
    Hongcai He\textsuperscript{\rm 1},
    Paul Weng\textsuperscript{\rm 3},
    Jie Shao\textsuperscript{\rm 1,2}\thanks{Corresponding author: Jie Shao.}
} \affiliations{
    \textsuperscript{\rm 1}University of Electronic Science and Technology of China, Chengdu, China\\
    \textsuperscript{\rm 2}Sichuan Artificial Intelligence Research Institute, Yibin, China\\
    \textsuperscript{\rm 3}Data Science Research Center, Duke Kunshan University, Kunshan, China\\
    \{sldeng, hehongcai\}@std.uestc.edu.cn, \{ztzheng, shaojie\}@uestc.edu.cn, paul.weng@dukekunshan.edu.cn
}

\begin{document}

\maketitle

\begin{abstract}
A major challenge in Reinforcement Learning (RL) is the difficulty
of learning an optimal policy from sparse rewards. Prior works
enhance online RL with conventional Imitation Learning (IL) via a
handcrafted auxiliary objective, at the cost of restricting the RL
policy to be sub-optimal when the offline data is generated by a
non-expert policy. Instead, to better leverage valuable information
in offline data, we develop Generalized Imitation Learning from
Demonstration (GILD), which meta-learns an objective that distills
knowledge from offline data and instills intrinsic motivation
towards the optimal policy. Distinct from prior works that are
exclusive to a specific RL algorithm, GILD is a flexible module
intended for diverse vanilla off-policy RL algorithms. In addition,
GILD introduces no domain-specific hyperparameter and minimal
increase in computational cost. In four challenging MuJoCo tasks
with sparse rewards, we show that three RL algorithms enhanced with
GILD significantly outperform state-of-the-art methods.
\end{abstract}

%

\section{Introduction}
\label{sec:introduction}

Reinforcement Learning (RL), which learns through trial and error
experience to maximize the cumulative reward, has achieved great
success in various dense reward tasks
\cite{DBLP:conf/icml/00010IZ23, DBLP:conf/nips/0001MDL23}. However,
RL agents still struggle to learn the optimal policy from real-world
scenarios with sparse rewards. For instance, there might be a reward
only if a navigation robot reaches the goal, with no reward feedback
on the numerous intermediate steps taken to arrive.

\begin{figure}[t]
    \centering
    \includegraphics[width=0.72\linewidth]{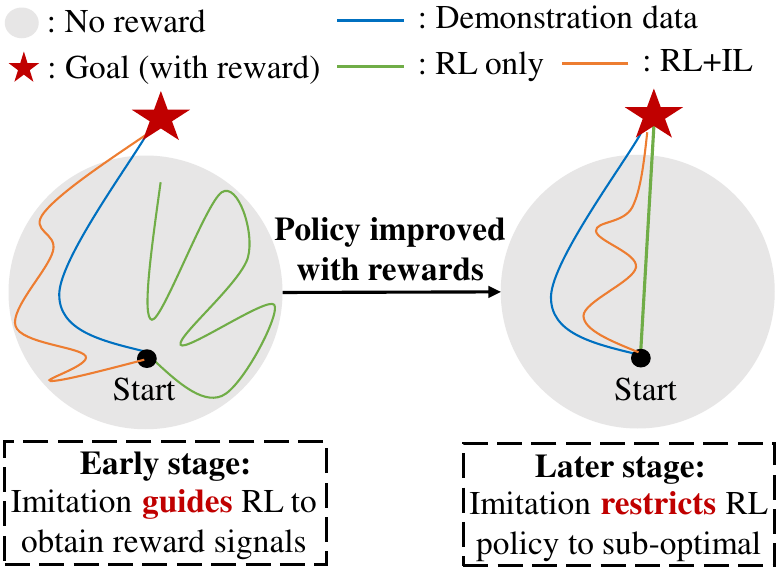}
    \caption{Illustration of RL+IL with sparse rewards. Conventional IL guides RL
to obtain reward signals in early stage (left), while restricting RL
policy to be sub-optimal in later stage (right).}
    \label{fig:RL+IL}
\end{figure}

To address the challenge of sparse rewards, prior works improve
online RL with conventional Imitation Learning (IL) by guiding the
agent to acquire reward signals that are essential for policy
improvement \cite{DBLP:conf/nips/Mendonca0KALF19,
DBLP:conf/nips/FujimotoG21, DBLP:conf/nips/RengarajanCKKS22}. These
RL+IL methods augment RL with conventional IL via a handcrafted
auxiliary objective, which constrains the agent to stay close to
behaviors observed in offline demonstration data. However, striking
a balance between RL and IL remains intractable, especially when the
agent is fed with sub-optimal demonstrations generated by humans. As
shown in Figure~\ref{fig:RL+IL}, conventional IL guides the agent to
obtain reward signals in early stage, but restricts the learned
policy to be sub-optimal in later stage. This observation leads to
the following research question: \textit{Is it possible to leverage
sub-optimal offline demonstrations for viable online RL with sparse
rewards, while not restricting the policy to be sub-optimal?} A
natural answer is to manually control or decay the influence of
imitation on policy optimization with some pre-defined schedule, but
at the cost of either spending massive time on hyperparameter tuning
or being exclusive to a specific RL algorithm
\cite{DBLP:conf/nips/FujimotoG21, DBLP:conf/nips/RengarajanCKKS22,
DBLP:conf/iclr/RengarajanVSKS22}.

By contrast, our key insight is to enhance online RL with a
meta-learned objective that leverages valuable information in
sub-optimal offline demonstrations, instead of RL with a handcrafted
objective in conventional IL. To achieve this, we develop
Generalized Imitation Learning from Demonstration (GILD), a flexible
module intended for diverse vanilla off-policy RL algorithms. We
devise a novel bi-level optimization framework for RL algorithms
enhanced with GILD, with meta-optimization of GILD at the upper
level and meta-training of RL at the lower level supported by the
meta-learned objective. We select off-policy RL as the vanilla
algorithm due to its superior sample efficiency compared with
on-policy alternatives. The advantage of sample efficiency extends
to meta-optimization, which updates GILD such that the policy
learned with RL+GILD is superior to that with RL+IL. We emphasize
that, in contrast to prior works that either augment RL with a
handcrafted IL objective or are exclusive to a specific RL
algorithm, GILD meta-learns a general IL objective and is intended
for diverse vanilla off-policy RL algorithms.

Our main results are as follows:
\begin{compactenum}[i.]
    \item GILD meta-learns a general IL objective to enhance online RL
via distilling knowledge from offline demonstrations, rather than
relying on a handcrafted IL objective in conventional IL. To the
best of our knowledge, GILD is the first to meta-learn an objective
to deal with sparse rewards.
    \item We integrate GILD with three vanilla off-policy RL algorithms (DDPG
\cite{DBLP:journals/corr/LillicrapHPHETS15}, TD3
\cite{DBLP:conf/icml/FujimotoHM18}, and SAC
\cite{DBLP:conf/icml/HaarnojaZAL18}) and evaluate them on four
challenging MuJoCo tasks with sparse rewards. Extensive experiments
show that the RL+GILD methods not only outperform the vanilla RL
methods and the conventional RL+IL variants, but also attain
asymptotic performance to the optimal policy.
    \item To further analyze the impact of GILD, we present several visualizations
including trajectories in a goal-reaching task and parameter
optimization paths in the MuJoCo tasks. These visualizations
demonstrate the aptitude of GILD at distilling knowledge from
sub-optimal demonstrations and instilling intrinsic motivation that
guides the RL agent towards the optimal policy.
    \item Finally, we observe that GILD converges exceptionally fast, making
it feasible to utilize RL+GILD at a few warm-start (e.g., $1\%$ of
total) time steps and subsequently drop GILD (RL only) to speed up
training. This highlights the potential to enhance RL with minimal
computational cost while achieving significant improvement.
\end{compactenum}

\section{Related Work}

Our work is mainly related to RL+IL, single-task meta-RL and
objective learning, which we discuss below. The closest methods to
our approach are LOGO \cite{DBLP:conf/iclr/RengarajanVSKS22} (RL+IL)
and Meta-Critic \cite{DBLP:conf/nips/ZhouLYWH20} (objective
learning).

\textbf{RL+IL.} We focus on reinforcement learning enhanced with
imitation learning (RL+IL) under sparse rewards, with the key idea
of utilizing demonstrations to assist policy learning. Prior works
have sought to (i) explicitly imitate behavior with demonstrations
to accelerate standard RL learning
\cite{DBLP:conf/nips/Mendonca0KALF19, DBLP:conf/nips/FujimotoG21} or
guide the RL agent towards non-zero reward regions of state-action
spaces \cite{DBLP:conf/nips/RengarajanCKKS22}, (ii) distill the
information within the demonstrations into an implicit prior
\cite{DBLP:conf/iclr/SinghLZYRL21, DBLP:conf/iclr/Hakhamaneshi0ZA22}
or combine multiple explicit and implicit priors obtained from
demonstrations \cite{DBLP:conf/nips/YanSW22}, and (iii) obtain
guidance from implicit imitation via aligning with the behavior
policy measured by KL-divergence
\cite{DBLP:conf/iclr/RengarajanVSKS22}. These methods strike a
balance between RL and IL at the cost of either spending massive
time on hyperparameter tuning or being exclusive to a specific RL
algorithm. Distinct from prior works, we propose a flexible module
named GILD, which is intended for diverse vanilla online RL
algorithms, to distill knowledge from offline demonstrations with a
meta-learned objective.

\textbf{Single-task meta-RL.} With the aim of accelerating learning
or improving performance, single-task meta-RL can meta-learn various
RL components, including (i) discount factor in scalar form
\cite{DBLP:conf/nips/XuHS18} or vector form
\cite{DBLP:conf/iclr/YinYX23}, (ii) reward function as an additive
intrinsic reward from data collected by RL
\cite{DBLP:conf/nips/ZhengOS18} or as the entire rewards from human
preference data \cite{DBLP:conf/nips/LiuBD022} and (iii) weights for
training samples to achieve better task awareness in model-based RL
\cite{DBLP:conf/nips/YuanDJ023}. By contrast, our proposed GILD
meta-learns a general IL objective from offline demonstrations and
automatically strikes a balance between RL and IL.

\textbf{Objective learning.} Different from the aforementioned works
that employ only a common objective function, objective learning in
RL or supervised learning aims to learn an objective. The learned
objective function has been exploited to (i) provide guidance for
accelerate learning in standard RL \cite{DBLP:conf/icml/XuRDLF19,
DBLP:conf/nips/XuHHOSS20, DBLP:conf/nips/ZhouLYWH20}, (ii) teach the
training of a student RL model \cite{DBLP:conf/nips/WuTXFQLL18,
DBLP:conf/iclr/FanTQ0L18, DBLP:conf/icml/HuangZTMSGS19,
DBLP:conf/nips/HaiPLLY23}, and (iii) improve generalization or
robustness to novel tasks with different dynamics
\cite{DBLP:conf/iccv/BaikCKCML21, DBLP:conf/nips/JinLRCWL23,
DBLP:conf/nips/NeymanR23}. Replacing conventional IL objective, our
approach enhances online RL with a meta-learned objective from
offline demonstrations.

\section{Preliminaries}
\label{sec:preliminaries}

\textbf{Standard RL.} Reinforcement learning typically considers an
infinite horizon Markov Decision Process (MDP), which is represented
as a tuple $<\mathcal{S}, \mathcal{A}, \mathcal{R}, \mathcal{P},
\gamma>$, with state space $\mathcal{S}$, action space
$\mathcal{A}$, reward function $\mathcal{R}$, transition dynamics
$\mathcal{P}$, and discount factor $\gamma$. At each timestep, given
state $s \in \mathcal{S}$, an RL agent takes action $a \in
\mathcal{A}$ based on its policy $\phi$, and receives reward
$r=\mathcal{R}(s, a)$ and new state $s'$ following the transition
dynamics $p(s'|s,a) \in \mathcal{P}$. The objective function of
policy $\phi$, known as the expected return, is defined as
$\mathcal{L}^{RL}(\phi)= -\mathbb{E}_{s \sim p,a \sim
\phi}[\sum_{t=0}^{\infty} \gamma^{t} r_{t}]$. With a bit abuse of
notation, we use $\phi$ to refer to both stochastic and
deterministic policy, as GILD is proposed for RL algorithms with
both stochastic policy (SAC) and deterministic policy (DDPG and
TD3).

Off-policy RL usually measures the objective with an actor-critic
architecture for superior sample efficiency via reusing past
experience $(s, a, r, s')$ stored in the replay buffer
$\mathcal{D}$. The critic parameterized by $\theta$, learns an
action-value function, which is defined as
$Q_{\theta}(s,a)=\mathbb{E}[\sum_{t=0}^{\infty} \gamma^{t} r_{t+1} |
s_{0}=s, a_{0}=a]$, to evaluate the expected return following policy
$\phi$ starting from state $s$ and action $a$. The critic is
updated to minimize the Mean-Square Bellman Error (MSBE) function:
\begin{equation} \small
   \begin{aligned}
    &\theta^{*} = \arg \min_{\theta}
    \mathcal{L}^{\mathrm{MSBE}}(\theta)\\
    &= \arg \min_{\theta} \mathbb{E}_{(s,a,r,s') \sim \mathcal{D}} \Big[Q_{\theta}(s,a)-\Big(r+\gamma Q_{\theta} \big(s',\phi(s') \big) \Big) \Big].
    \label{eq: critic}
    \end{aligned}
\end{equation}

The policy $\phi$, known as the actor, is updated to minimize
the loss given by the critic:
\begin{equation} \small
    \phi^{*}=\arg \min_{\phi} \mathcal{L}^{\mathrm{RL}}_{\theta}(\phi) = \arg \min_{\phi} \mathop{\mathbb{E}}\limits_{s \sim \mathcal{D}} \Big[-Q_{\theta} \big(s,\phi(s) \big) \Big].
\end{equation}

\textbf{RL+IL.} The most commonly used form of IL is Behaviour
Cloning (BC), which focuses on imitating behaviors in demonstration
data $\mathcal{D}^{\mathrm{dem}}$ using supervised learning. The
supervised learning objective for it is defined as
$\mathcal{L}^{\mathrm{IL}}(\phi)= N^{-1} \sum_{(s,a) \in
\mathcal{D}^{\mathrm{dem}}}(\phi(s)-a)^{2}$ for the deterministic
policy and $\mathcal{L}^{\mathrm{IL}}(\phi)= -N^{-1}\sum_{(s,a) \in
\mathcal{D}^{\mathrm{dem}}}{\log(\pi_{\phi}(a|s))}$ for the
stochastic policy. Recent online RL approaches
\cite{DBLP:conf/nips/Mendonca0KALF19, DBLP:conf/nips/FujimotoG21,
DBLP:conf/nips/RengarajanCKKS22} utilize IL as an auxiliary
objective added to the update steps of an RL policy, to push the
policy towards behaviors in demonstrations:
\begin{equation} \small
    {\phi^{*}}= \arg \min_{\phi} \big(\mathrm{w_{rl}} \mathcal{L}^{RL}(\phi) + \mathrm{w_{il}} \mathcal{L}^{IL}(\phi) \big),
    \label{eq:RL+IL}
\end{equation}
where $\mathrm{w_{rl}}$ and $\mathrm{w_{il}}$ are hyperparameters
that control the influence of RL and IL on policy optimization.

\section{Methodology}
\label{sec:methodolocy}

In this section, we present off-policy RL augmented by GILD, which
is formalized as a bi-level optimization framework, with (i)
meta-optimization of GILD at the upper level and (ii) meta-training
of RL at the lower level supported by the meta-learned objective.
Following notations in off-policy RL and meta-RL, we denote the
parameters of actor, critic, and GILD network as $\phi$, $\theta$,
and $\omega$ respectively. We denote the objective learned by GILD
$\omega$ as $\mathcal{L}^{\mathrm{GILD}}_{\omega}(\phi)$, whose
input depends on actor parameter $\phi$.

\begin{algorithm}[t] \small
\caption{RL+GILD} \label{alg:RL+GILD} \textbf{Input}: Actor $\phi$,
critic $\theta$, GILD $\omega$, demonstration data
$\mathcal{D}^{\mathrm{dem}}$, and empty replay buffer $\mathcal{D}$
\begin{algorithmic}[1]
    \WHILE{not converging}
        \STATE Collect data from the environment and store in $\mathcal{D}$;
        \STATE \textbf{meta-training:}
        \STATE Sample $(s,a,r,s')$ from $\mathcal{D}$, and $(s^{\mathrm{d}}, a^{\mathrm{d}})$ from $\mathcal{D}^{\mathrm{dem}}$;
        \STATE Update critic $\theta$ via Eq.~\eqref{eq:update: critic};
        \STATE Pseudo-update actor $\hat{\phi}$ with RL+IL via Eq.~\eqref{eq:update: actor: RL+IL};
        \STATE Update actor $\phi$ with RL + GILD via Eq.~\eqref{eq:update: actor: RL+GILD};
        \STATE \textbf{meta-optimization:}
        \STATE Update GILD $\omega$ via Eq.~\eqref{eq:update: GILD};
    \ENDWHILE
\end{algorithmic}
\end{algorithm}

\subsection{Overview}

\begin{figure}[t]
\centering
\includegraphics[width=0.72\linewidth]{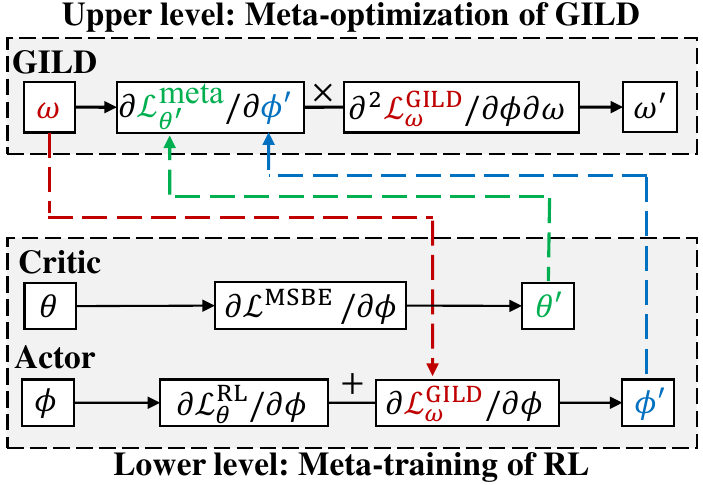}
\caption{Workflow of the bi-level optimization framework, with
meta-optimization of GILD at the upper level and meta-training of RL
at the lower level supported by
$\mathcal{L}^{\mathrm{GILD}}_{\omega}$.} \label{fig:Workflow}
\end{figure}

The proposed GILD aims to enhance online RL with a meta-learned
objective $\mathcal{L}^{\mathrm{GILD}}_{\omega}(\phi)$ that distills
knowledge from sub-optimal offline demonstrations, rather than
relying on conventional IL via supervised learning. More specially,
GILD is updated with meta-loss
$\mathcal{L}^{\mathrm{meta}}_{\theta}(\phi)$, which optimizes GILD
in the direction that the policy learned with RL+GILD is superior to
policy with RL+IL. See Algorithm~\ref{alg:RL+GILD} for a pseudocode
of bi-level paradigm and Figure~\ref{fig:Workflow} for a workflow of
bi-level optimization. The overall objective is formulated as:
\begin{align} \small
    \min_{\omega} \quad & \mathcal{L}^{\mathrm{meta}}_{\theta^{*}} ( \phi^{*}), \notag \\
    \text {\textit{s.t.}} \quad &
    \left\{ \begin{aligned}
    \phi^{*} &= \arg \min_{\phi} \big( \mathcal{L}^{\mathrm{RL}}_{\theta^{*}}(\phi)+\mathcal{L}^{\mathrm{GILD}}_{\omega}(\phi) \big), \\
    \theta^{*} &= \arg \min_{\theta} \big( \mathcal{L}^{\mathrm{MSBE}}(\theta ) \big),
    \end{aligned}\right.
    \label{eq:overview}
\end{align}
where meta-training at the lower level includes conventional critic
learning and policy learning supported by GILD. Thanks to
meta-optimization of GILD at the upper level, the policy learned
with RL+GILD could be superior to the policy learned with RL+IL in
Eq.~\eqref{eq:RL+IL}. This bi-level optimization enables GILD to
distill knowledge from offline data and instills in the online RL
agent the intrinsic motivation towards optimal policy, hence not
restricting RL policy to be sub-optimal.

\subsection{General Imitation Learning Objective}

As previously discussed, a policy trained using non-expert
demonstrations via Eq.~\eqref{eq:RL+IL} is restricted to be
sub-optimal. We address this issue with general imitation learning
objective, which enhances RL by leveraging valuable information in
sub-optimal demonstrations $\mathcal{D}^{\mathrm{dem}}$. The
supervised learning objective function for IL can be formalized as
$\mathcal{L}^{\mathrm{IL}}(\phi)=f(\phi;
\mathcal{D}^{\mathrm{dem}})$, with a handcrafted loss function
$f(\cdot)$ (e.g., mean square error), which restricts agent around
the behavior policy. We devise GILD as a neural network parametrized
by $\omega$ to meta-learn a general update function
$f_{\omega}(\cdot)$, which produces a general IL objective
$\mathcal{L}^{\mathrm{GILD}}_{\omega}(\phi)=f_{\omega}(\phi;
\mathcal{D}^{\mathrm{dem}})$.

We implement GILD as a three-layer fully connected network for the
following considerations: (i) GILD should be flexible to be
integrated with diverse vanilla off-policy RL algorithms; (ii) For
the feasibility to be applied to downstream tasks (e.g., use
convolutional neural networks as GILD's backbone for image-based
autonomous driving task), GILD ought to introduce no domain-specific
hyperparameter; (iii) GILD is supposed to enhance off-policy RL
without reducing the superior sample efficiency.

\textbf{Building connection between lower-level and upper-level.}
(i) Upper-to-lower: To update the RL policy, the general IL
objective $\mathcal{L}^{\mathrm{GILD}}_{\omega}(\cdot)$ outputted by
GILD must be differentiable w.r.t. policy parameter $\phi$, which
means the input of GILD should depend on the actor. This is
satisfied in an end-to-end manner: GILD takes the combination of
demonstration state-action pair $(s^{\mathrm{d}}, a^{\mathrm{d}})$
and actor's action $a=\phi(s^{\mathrm{d}})$ as the input. (ii)
Lower-to-upper: To update GILD, the meta-loss, which is the
action-value function $Q_{\theta}(\cdot)$ for sample efficiency
consideration, must be differentiable w.r.t. GILD parameter
$\omega$. As depicted in Figure~\ref{fig:Workflow}, the connection
between $\theta$ and $\omega$ is built as follows. First, $\theta$
is differentiable w.r.t. $\phi$ since $Q_{\theta} \big(s,\phi(s)
\big)$ takes action $\phi(s)$ as the input. Second, $\phi$ is
differentiable w.r.t. $\omega$ since it is updated with
$\mathcal{L}^{\mathrm{GILD}}_{\omega}(\phi)$. Therefore, $\theta$ is
differentiable w.r.t. $\omega$.

\subsection{Bi-Level Optimization}
\label{sec:bi-level optimization}

After defining general IL objective
$\mathcal{L}^{\mathrm{GILD}}_{\omega}(\phi)$ and building a
connection for bi-level optimization, we divide the bi-level
objective in Eq.~\eqref{eq:overview} into meta-training
(lower-level) and meta-optimization (upper-level) to solve them
respectively. Note that we omit tricks (e.g., target network and
entropy regularizer) used in different off-policy algorithms here
for simplicity. Detailed algorithms for three vanilla off-policy RL
algorithms enhanced with GILD are presented in the supplementary
material.

\textbf{Lower-level: meta-training.} After collecting a set
$\mathcal{D}$ of transitions $(s, a, r, s')$ through interacting
with the environment, off-policy RL reuses these past experiences to
update critic and actor sequentially. The critic is updated with a
batch of $N$ transitions to minimize the MSBE function as:
\begin{equation} \small
    \begin{aligned}
    \theta^{(k+1)} = \theta^{(k)} - \alpha \nabla_{\theta} \frac{1}{N}\sum_{(s,a,r,s') \sim \mathcal{D}} \Big[ Q_{\theta}(s,a)-\\
    \Big(r+\gamma Q_{\theta} \big(s',\phi(s') \big) \Big) \Big]^{2} \Big|_{\theta^{(k)}, \phi^{(k)}},
    \label{eq:update: critic}
    \end{aligned}
\end{equation}
where $\theta^{(k+1)}$ denotes the updated parameter $\theta^{(k)}$
at step $k$, $\alpha$ is the learning rate, and $\gamma$ is the
discount factor.

Before updating the actor, we \textit{pseudo-update} the actor with
RL+IL. The pseudo-updated actor is intended for computing the
meta-loss later, which guides the policy learned with RL+GILD to be
potentially superior to that with RL+IL. Pseudo-update means that we
do not directly update actor $\phi^{(k)}$, but update a copy of the
current actor $\hat{\phi}^{(k)}$:
\begin{equation} \small
    \begin{aligned}
   \hat{\phi}^{(k+1)} = \hat{\phi}^{(k)} - \alpha \nabla_{\hat{\phi}} \Big[&\mathrm{w_{rl}} \frac{1}{N} \sum_{(s,a) \sim \mathcal{D}} -Q_{\theta} \big(s,\hat{\phi}(s) \big) + \\
   &\mathrm{w_{il}} \mathcal{L}^{IL}(\hat{\phi}) \Big] \Big|_{\theta^{(k+1)}, \hat{\phi}^{(k)}},
    \label{eq:update: actor: RL+IL}
    \end{aligned}
\end{equation}
where $\alpha$ is the learning rate, $\mathcal{L}^{IL}(\hat{\phi})$
is the conventional IL objective used in Eq.~\eqref{eq:RL+IL}, and
$\mathrm{w_{rl}}$ and $\mathrm{w_{il}}$ are hyperparameters that
control the influence of RL and IL on policy optimization. Following
TD3+BC \cite{DBLP:conf/nips/FujimotoG21}, an approach for off-policy
RL+IL, we assign the hyperparameters as $\mathrm{w_{rl}} =
{\beta}/{\frac{1}{N} \sum_{s,a}|Q_{\theta}(s,a)|}$ and
$\mathrm{w_{il}}=1$ for off-policy RL+IL baselines in our
experiment, with $\beta$=2.5 provided by the authors. Following
EMRLD \cite{DBLP:conf/nips/RengarajanCKKS22}, an approach for
on-policy RL+IL, we set $\mathrm{w_{rl}}=1$ and $\mathrm{w_{il}}=1$
for on-policy RL+IL baselines.

After the pseudo-update, the actor is updated to minimize both
objectives given by the critic and GILD:
\begin{equation} \small
    \begin{aligned}
    \phi^{(k+1)} = \phi^{(k)} - \alpha \nabla_{\phi} \Big[&\frac{1}{N} \sum_{(s,a) \sim \mathcal{D}} -Q_{\theta} \big( s,\phi(s) \big) +\\
    &\mathcal{L}^{GILD}_{\omega}(\phi) \Big] \Big|_{\theta^{(k+1)}, \phi^{(k)}, \omega^{(k)}},
    \label{eq:update: actor: RL+GILD}
    \end{aligned}
\end{equation}
where $\mathcal{L}^{\mathrm{GILD}}_{\omega}(\phi)=N^{-1} \sum
f_{\omega} \big(s^{\mathrm{d}}, a^{\mathrm{d}}, \phi(s^{\mathrm{d}})
\big)$ with a batch of $N$ state-action pairs $(s^{\mathrm{d}},
a^{\mathrm{d}})$ sampled from demonstrations
$\mathcal{D}^{\mathrm{dem}}$.

\textbf{Upper-level: meta-optimization.} The intuition of the
meta-loss is to update GILD $\omega$ in the direction that the
policy learned with RL+GILD is superior to that with RL+IL. This
superiority could be measured quantitatively by the difference in
action-value function $Q_{\theta}(\cdot)$ as:
\begin{equation} \small
    \begin{aligned}
    \mathcal{L}^{\mathrm{meta}}_{\theta} (\phi)= \frac{1}{N} \sum_{s^{\mathrm{val}} \sim \mathcal{D}} \Big[ \tanh \Big( Q_{\theta}\big( s^{\mathrm{val}}, \phi(s^{\mathrm{val}}) \big) -\\
    Q_{\theta} \big( s^{\mathrm{val}}, \hat{\phi}(s^{\mathrm{val}}) \big) \Big) \Big] \Big|_{\theta^{(k+1)}, \phi^{(k+1)}, \hat{\phi}^{(k+1)}, \omega^{(k)}},
    \label{eq:meta-loss}
    \end{aligned}
\end{equation}
where $s^{\mathrm{val}}$ is the validation states sampled from past
experiences for sample efficiency consideration, $\phi$ is from
performing Eq.~\eqref{eq:update: actor: RL+GILD}, and $\hat{\phi}$
is from performing Eq.~\eqref{eq:update: actor: RL+IL}. The
derivative of $\mathcal{L}^{\mathrm{meta}}_{\theta} (\phi)$ w.r.t.
$\omega$ is calculated using the chain rule:
\begin{equation} \small
    \begin{aligned}
    \frac{\partial \mathcal{L}^{\mathrm{meta}}_{\theta} ({\phi}) } {\partial \omega}
    &= \frac{\partial \mathcal{L}^{\mathrm{meta}}_{\theta}({\phi}) } {\partial \phi} \cdot \frac{\partial \phi} {\partial \omega} \Big|_{\theta^{(k+1)}, \phi^{(k+1)}, \omega^{(k)}}\\
    &= \frac{\partial \mathcal{L}^{\mathrm{meta}}_{\theta}({\phi}) } {\partial \phi} \cdot
    g^{(k)}_{\omega} \Big|_{\theta^{(k+1)}, \phi^{(k+1)}, \omega^{(k)}},
    \label{eq:derivative: meta-loss total}
    \end{aligned}
\end{equation}
where $g^{(k)}_{\omega}$, which is actually the second-order
derivative, can be obtained as follows. Since $Q(\cdot)$ in
Eq.~\eqref{eq:update: actor: RL+GILD} is a constant $c$ that is
independent of $\omega$, we simplify $\phi^{(k+1)}$ and
$g^{(k)}_{\omega}$ as:
\begin{equation} \small
    \begin{aligned}
    \phi^{(k+1)} = \phi^{(k)} - \alpha \frac{\partial \mathcal{L}^{\mathrm{GILD}}_{\omega}(\phi)}{\partial \phi}  + c\ \Big|_{\phi^{(k)}, \omega^{(k)}},\\
    g^{(k)}_{\omega} = \frac{\partial \phi^{(k+1)}} {\partial \omega}
    = -\alpha \frac{\partial^{2} \mathcal{L}^{\mathrm{GILD}}_{\omega}(\phi)}{\partial \phi \partial\omega} \Big|_{\phi^{(k)}, \omega^{(k)}}.
    \label{eq:derivative: meta-loss second-order}
    \end{aligned}
\end{equation}
Combining Eq.~\eqref{eq:derivative: meta-loss total} and
Eq.~\eqref{eq:derivative: meta-loss second-order}, we get the
derivative w.r.t. $\omega$. Then, $\omega$ is meta-optimized as:
\begin{equation} \small
    \begin{aligned}
    &\omega^{(k+1)} = \omega^{(k)} + \\ &\alpha^{2} \frac{\partial \mathcal{L}^{\mathrm{meta}}_{\theta}({\phi}) } {\partial \phi}\Big|_{\phi^{(k+1)}, \omega^{(k)}}  \cdot
     \frac{\partial^{2} \mathcal{L}^{\mathrm{GILD}}_{\omega}(\phi)}{\partial \phi \partial\omega} \Big|_{\phi^{(k)} \omega^{(k)}}.
     \label{eq:update: GILD}
    \end{aligned}
\end{equation}

\section{Experiments}
\label{sec:experiments}

\begin{table*}[t] 
  \centering \resizebox{0.72\linewidth}{!}{
  \begin{tabular}{lccccc}
    \toprule
    Algorithm   & Hopper-v2    & Walker2d-v2 & HalfCheetah-v2   & Ant-v2   & Point2D Navigation\\
    \midrule
    DDPG        & 2122.9$\pm$590.7  & 1519.4$\pm$881.8  & 3349.1$\pm$1489.6  & 339.0$\pm$109.2  & 19.7$\pm$13.3 \\
    DDPG+IL     & 2378.4$\pm$906.1  & 1867.8$\pm$489.5  & 5603.9$\pm$1129.9  & 575.1$\pm$215.4 & 47.5$\pm$16.8 \\
    DDPG+GILD (ours)    & \underline{2804.0$\pm$235.4}  & \underline{2632.1$\pm$373.0}  & \underline{9987.7$\pm$511.9}  & \underline{971.6$\pm$296.7}  & \underline{71.0$\pm$8.7} \\
    \midrule
    TD3         & 1320.8$\pm$413.9  & 1426.6$\pm$1413.0  & 3251.3$\pm$1135.4  & 1712.3$\pm$562.8   & 24.0$\pm$10.7 \\
    TD3+IL      & 2437.9$\pm$890.2  & 2488.5$\pm$903.7  & 5843.9$\pm$1321.0  & 2660.2$\pm$395.5  & 55.8$\pm$13.8 \\
    TD3+GILD (ours)     & \underline{\textbf{3538.6$\pm$104.6}}  & \underline{4113.6$\pm$280.5}  & \underline{9997.6$\pm$754.9}  & \underline{4864.6$\pm$699.1}  & \underline{75.1$\pm$9.7} \\
    \midrule
    SAC         & 2235.1$\pm$569.6  & 1643.2$\pm$809.5  & 3946.2$\pm$485.2  & 2106.8$\pm$718.0  & 43.6$\pm$17.2 \\
    SAC+IL      & 2989.6$\pm$263.3  & 3102.1$\pm$476.5  & 6503.2$\pm$802.5  & 3370.8$\pm$466.3  & 67.1$\pm$14.4 \\
    SAC+GILD (ours)     & \underline{3470.6$\pm$85.2}  & \underline{\textbf{4840.4$\pm$243.8}}  & \underline{\textbf{11161.5$\pm$552.6}}  & \underline{\textbf{5335.3$\pm$246.9}}  & \underline{\textbf{79.8$\pm$6.5}} \\
    \midrule
    PPO         & 1332.5$\pm$1356.33  & 6.3$\pm$13.3  & -10.3$\pm$514.2  & 637.6$\pm$191.3  & 23.6$\pm$15.5 \\
    PPO+IL      & 1831.7$\pm$279.8  & 2649.5$\pm$86.8  & 2781.3$\pm$61.7  & 1759.8$\pm$7.7  & 43.2$\pm$8.6 \\
    LOGO        & \underline{3465.80$\pm$88.2}  & \underline{4537.5$\pm$293.4}  & 5264.0$\pm$486.5  & \underline{4589.5$\pm$992.9}  & \underline{77.8$\pm$8.0} \\
    Meta-Critic     & 3185.2$\pm$526.9  & 3807.0$\pm$1377.1  & \underline{6811.6$\pm$3981.0}  & 1588.9$\pm$782.8  & 68.6$\pm$23.7 \\
    DiffAIL     & 2494.3$\pm$77.5  & 2848.3$\pm$153.4  & 5978.6$\pm$237.0  & 3650.1$\pm$183.9  & 51.3$\pm$4.1 \\
    \bottomrule
  \end{tabular}
  }
  \caption{Comparison on max average return of three vanilla off-policy RL
algorithms, RL+IL and RL+GILD, along with (on-policy or
state-of-the-art) methods. Results are run on sparse environments
over 5 trials, and ``$\pm$'' captures the standard deviation over
trials. Max value for each category is underlined, and max value
overall is in bold.}
  \label{tab:max-avg-return summary}
\end{table*}

\textbf{Research questions.} Our experiments are designed to
investigate the following research questions:
\begin{compactitem}
    \item \textbf{RQ1}: What is the enhancement of RL+GILD compared with RL+IL and objective learning methods?
    \item \textbf{RQ2}: How does GILD enhance RL compared with conventional IL?
    \item \textbf{RQ3}: What are the effects of different meta-loss designs and warm-start steps for GILD?
    \item \textbf{RQ4}: How to mitigate the computational cost of GILD?
\end{compactitem}

\textbf{Benchmarks and vanilla RL algorithms.} We conduct
experiments on four challenging MuJoCo tasks with sparse rewards.
Following EMRLD \cite{DBLP:conf/nips/RengarajanCKKS22}, the agent
gets a reward only after it has moved a certain number of units
along the correct direction, making the rewards sparse. We take
three popular off-policy RL algorithms as our vanilla algorithms,
which are DDPG \cite{DBLP:journals/corr/LillicrapHPHETS15}, TD3
\cite{DBLP:conf/icml/FujimotoHM18}, and SAC
\cite{DBLP:conf/icml/HaarnojaZAL18}. We use open-source
implementations of
``OurDDPG''\footnote{https://github.com/sfujim/TD3/blob/master/OurDDPG.py},
TD3\footnote{https://github.com/sfujim/TD3/blob/master/TD3.py}, and
SAC\footnote{https://github.com/pranz24/pytorch-soft-actor-critic}.

\textbf{Baselines.} In addition to the above three vanilla RL
algorithms and their RL+IL variants, we run the following
(state-of-the-art) algorithms using either author-provided or
open-source implementation: (i) \textbf{LOGO}: We re-run Learning
Online with Guidance Offline (LOGO)
\cite{DBLP:conf/iclr/RengarajanVSKS22}, which merges TRPO (on-policy
RL) with an additional policy step using sub-optimal demonstration
data. (ii) \textbf{Meta-Critic}: We re-run Meta-Critic
\cite{DBLP:conf/nips/ZhouLYWH20}, which meta-learns an additional
objective for off-policy RL. (iii) \textbf{DiffAIL}: We re-run
Diffusion Adversarial Imitation Learning (DiffAIL)
\cite{DBLP:conf/aaai/WangWPZY24}, which introduces the diffusion
model into adversarial IL. (iv) \textbf{PPO} and \textbf{PPO+IL}: We
re-run PPO \cite{DBLP:journals/corr/SchulmanWDRK17} and its RL+IL
variant to compare with on-policy RL. (v) \textbf{Expert} and
\textbf{Behavior}: Following LOGO
\cite{DBLP:conf/iclr/RengarajanVSKS22}, we train vanilla RL
algorithms in the dense reward environment to provide three Expert
baselines. We use the partially trained Expert that is still at a
sub-optimal stage as the Behavior baselines to provide demonstration
data for the corresponding RL+IL and RL+GILD algorithms.

\textbf{Implementation details.} To ensure a fair and identical
experimental evaluation across algorithms, we train the (RL+IL and
RL+GILD) variant using the same hyperparameters as their vanilla
algorithms and introduce no domain-specific parameters. We train
off-policy algorithms for 1 million steps with sparse rewards and
evaluate them every 5000 steps with dense rewards. On-policy
algorithms are trained with more steps (e.g., 30 million) to ensure
convergence. Results are averaged over five random seeds and the
standard deviation is shown with the shaded region or error bar. Our
code is available at \url{https://github.com/slDeng1003/GILD}.

\subsection{RQ1: Comparison w. RL+IL \& Objective Learning}

The max average returns for all methods are summarized in
Table~\ref{tab:max-avg-return summary}. We display the most
representative learning curve of vanilla TD3 algorithms with its
corresponding TD3+IL and TD3+GILD variants in
Figure~\ref{fig:Learning_curve+Avg_Norm_scores}, and more learning
curves are in the supplementary material. Besides,
Figure~\ref{fig:Learning_curve+Avg_Norm_scores} presents the average
normalized score of vanilla algorithms, their corresponding
variants, and Behavior algorithms. Scores are normalized using the
max average return of Expert (with a score of 100).

\begin{figure}[t]
    \centering
    \includegraphics[width=0.92\linewidth]{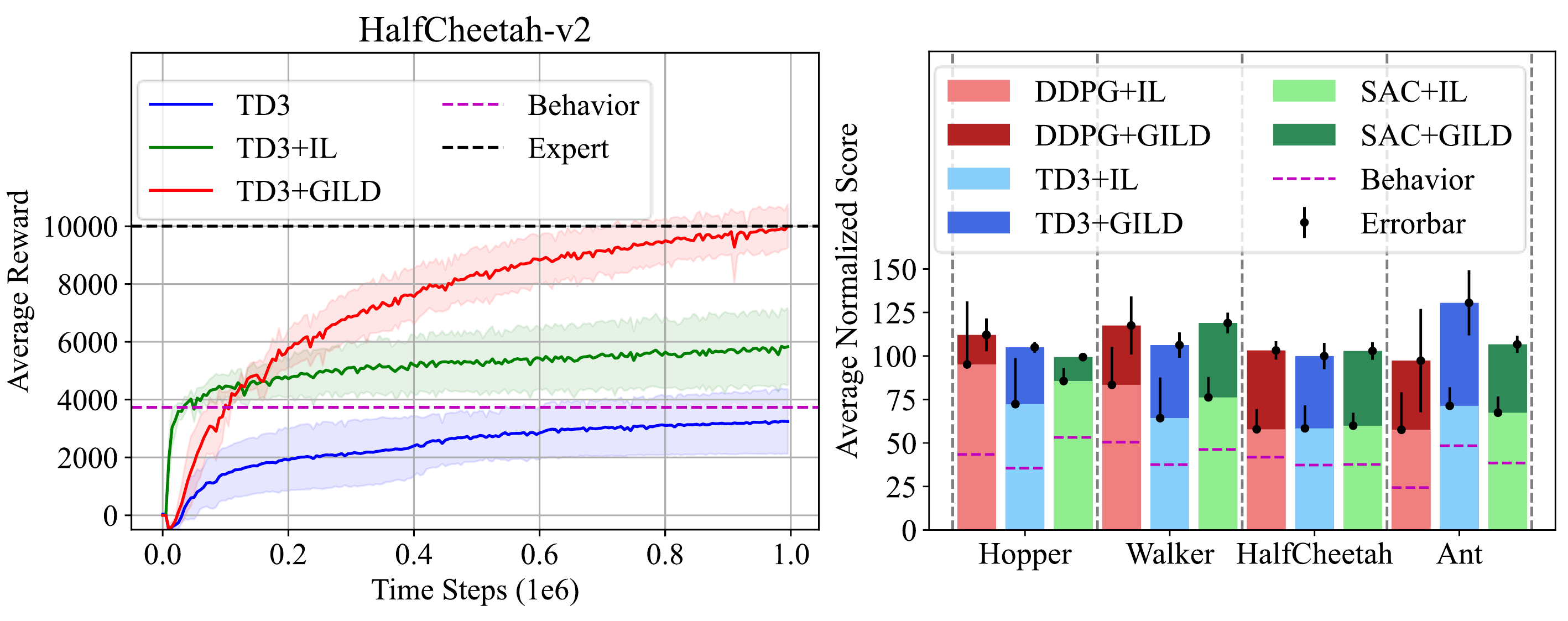}
\caption{Learning curve with mean-std (left) and average normalized
score (right) in the MuJoCo task(s) with sparse rewards. We
normalized the scores using max average return of Expert (with a
score of 100).}
    \label{fig:Learning_curve+Avg_Norm_scores}
\end{figure}

\begin{figure*}[t]
    \centering
    \includegraphics[width=0.96\linewidth]{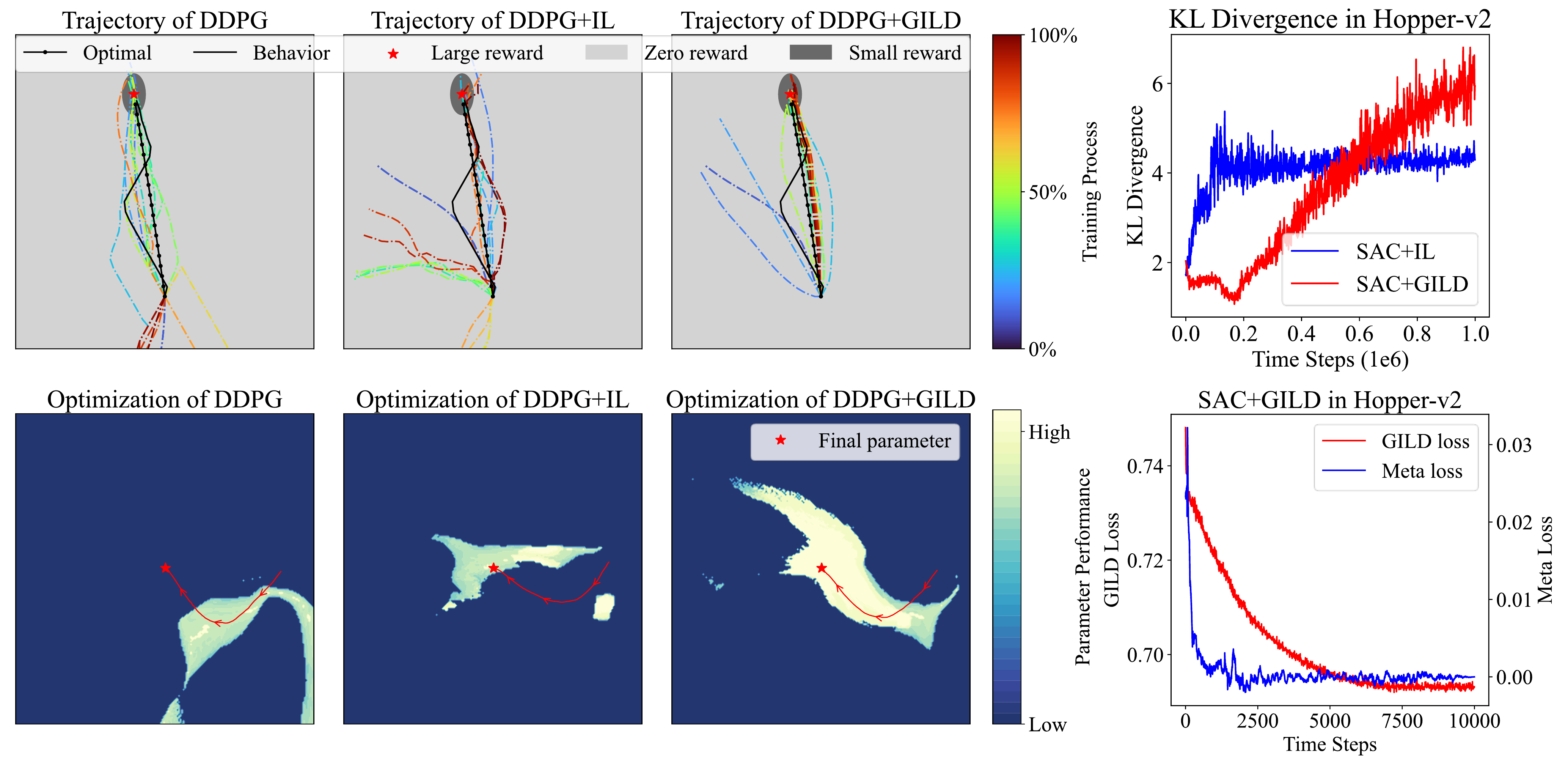}
    \caption{(i) Left: Visualization of evaluation trajectories and corresponding
policy optimization paths for DDPG, DDPG+IL, DDPG+GILD in Point2D
Navigation. The red star denotes the goal to reach, as well as
parameters for the final policy. (ii) Right: KL divergence and loss
analysis for SAC+IL and SAC+GILD.}
    \label{fig:Eval_traj+Opt_path+KL+Loss}
\end{figure*}

In all four benchmarks, our RL+GILD methods significantly outperform
the other baselines, while vanilla algorithms fail in most cases due
to the sparsity of reward. Learning curve of TD3+IL rises quickly in
the initial stage of learning, indicating the agent obtains non-zero
rewards via imitation, which underscores the necessity of imitating
demonstrations. However, the policy learned by TD3+IL is restricted
to be sub-optimal, while TD3+GILD smoothly surpasses the Behavior
policy and attain asymptotic or superior performance to the Expert
policy, emphasizing the benefit of leveraging insights from
sub-optimal demonstrations. LOGO exhibits comparable performance to
RL+GILD across several tasks, albeit with noticeably lower sample
efficiency and slower learning speed as shown in Table~\ref{tab:run
time}. Meta-Critic achieves commendable performance in a subset of
benchmarks, although it struggles to reach the Expert performance
due to its inability to utilize information in demonstrations.
DiffAIL does not attain good metrics because it relies heavily on
the quality of offline data collected by sub-optimal policy. More
results for RQ1 are in the supplementary material.

\subsection{RQ2: Visualization and Loss Analysis}
\label{sec:experiments-visualization}

To investigate how GILD enhances the vanilla RL algorithms compared
with conventional IL, we (i) visualize the evaluation trajectories
and corresponding optimization paths of DDPG, DDPG+IL and DDPG+GILD,
(ii) display the KL divergence of SAC+IL and SAC+GILD with the
Behavior policy, and (iii) plot the value of general IL objective
and meta-loss to demonstrate the convergence of GILD. Following
Meta-Critic \cite{DBLP:conf/nips/ZhouLYWH20}, curves are uniformly
smoothed for clarity. Further visualization results are in the
supplementary material.

\textbf{(i) Trajectory visualization:} We run DDPG, DDPG+IL and
DDPG+GILD in Point2D Navigation
\cite{DBLP:conf/nips/RengarajanCKKS22}, a 2-dimensional
goal-reaching environment with $|\mathcal{S}|$=2, $|\mathcal{A}|$=2.
We plot the trajectories of the on-learning model at each evaluation
at the top-left of Figure~\ref{fig:Eval_traj+Opt_path+KL+Loss}, and
different training periods serve as colors of each trajectory. On
the one hand, trajectories of DDPG+IL in the early stage are quite
similar to the Behavior trajectories, indicating that the agent
quickly learns a policy close to the Behavior policy via imitation.
However, trajectories of DDPG+IL in the later stage deviate to the
wrong direction towards the goal (red star), due to the incongruity
between RL and conventional IL. On the other hand, DDPG+GILD
eliminates the incongruity by leveraging the valuable information in
demonstrations, with trajectories consistently resemble the optimal
after the initial stage.

\textbf{(ii) Optimization path visualization:} Corresponding to the
aforementioned trajectories, we display policy optimization paths
(red line with arrow) in the parameter space at the bottom-left of
Figure~\ref{fig:Eval_traj+Opt_path+KL+Loss}. Following network
visualization in \citet{DBLP:conf/nips/Li0TSG18}, we apply principal
component analysis to reduce the dimension of policy parameter
$\phi$, and take the top-2 representative components for plotting on
the 2D surface. Every point on the surface represents a policy.
These policies are densely evaluated over 10 episodes to get the
average reward values, which serve as colors of the points. The
policy optimization paths demonstrate that DDPG+GILD moves directly
and quickly to the high reward area (brighter color) on the surface,
while the vanilla DDPG and DDPG+IL struggle to move beyond the low
reward area (darker color) and finally learn a sub-optimal or bad
policy.

\textbf{(iii) KL divergence analysis:} The stochastic policy in SAC
provides feasibility to calculate the KL divergence between the
learning policy and the Behavior policy. We display it at the
top-right of Figure~\ref{fig:Eval_traj+Opt_path+KL+Loss} and find
that policy learned by SAC+IL is constrained to be similar to
Behavior due to the handcrafted objective. By contrast, the policy
learned by SAC+GILD leverages knowledge distilled from
demonstrations and moves beyond the Behavior policy with
consistently rising KL divergence after the early stage.

\textbf{(iv) Loss analysis:} We plot values of general IL objective
$\mathcal{L}^{\mathrm{GILD}}_{\omega}$ and meta-loss
$\mathcal{L}^{\mathrm{meta}}_{\theta}$ at the bottom-right of
Figure~\ref{fig:Eval_traj+Opt_path+KL+Loss}, which demonstrates that
GILD converges exceptionally quickly (within $1\%$ of total steps)
under the supervision of meta-loss. Meta-loss drops rapidly around
zero after 1000 steps, verifying that GILD has distilled most of the
knowledge in demonstrations from $t_{\mathrm{ws}}\times B \div N
\approx 640$ times of processing each data, where
$t_{\mathrm{ws}}$=10000 is the warm-start steps, $B$=256 is the
batch size, and $N \approx 4000$ is the number of samples. As we
will discuss later in RQ3 and RQ4, GILD's rapid convergence
indicates that we can utilize GILD with a few warm-start (ws) steps
(e.g., $1\%$ of total steps) and subsequently drop GILD to speed up
training.

\subsection{RQ3: Ablation on Meta-Loss and Warm-Start}

\textbf{(i) Ablation on meta-loss design:} As discussed in
Methodology, $Q_{\theta}(\hat{\phi})$ in the meta-loss is
independent to GILD parameter $\omega$, so the most intuitive
meta-loss is defined as $\mathcal{L}^{\mathrm{meta}}_{\theta} (\phi)
= \mathbb{E}[Q_{\theta}(\phi)]$. This intuitive meta-loss aims to
maximize the performance of policy $\phi$ updated with GILD
$\omega$. We evaluate these two meta-loss designs in the most
challenging sparse Ant-v2 benchmark ($|\mathcal{S}|=111$,
$|\mathcal{A}|=8$) and report the max average return in
Table~\ref{tab:ablation on meta-loss}. We find that GILD with
meta-loss in Eq.~\eqref{eq:meta-loss} outperforms GILD with the
intuitive meta-loss, which also improves vanilla RL algorithms.

\begin{table}[t]
    \centering \resizebox{0.82\linewidth}{!}{
    \begin{tabular}{ccc}
    \toprule
    Algorithm & Meta-loss in Eq.~\eqref{eq:meta-loss} & Intuitive meta-loss \\
    \midrule
    DDPG+GILD & \textbf{971.6+-296.7}    & 883.1+-254.8 \\
    TD3+GILD  & \textbf{4864.6+-699.1}   & 4259.4+-716.3 \\
    SAC+GILD  & \textbf{5335.3+-246.9}   & 4851.0+-218.5 \\
    \bottomrule
    \end{tabular}
    }
    \caption{Ablation study on different designs of meta-loss applied to three
RL+GILD methods in the sparse Ant benchmark. Max value for each
method is in bold.}
    \label{tab:ablation on meta-loss}
\end{table}

\textbf{(ii) Ablation on GILD warm-start (ws):} As discussed in RQ2,
GILD converges quickly with a few warm-start ($1\%$ of total) steps.
To investigate the influence of warm-start on the performance, we
implement different warm-start steps on the RL+GILD methods.
Training of a policy learned by RL+GILD+$1\%$ws is split into two
training stages: (i) RL+GILD stage: during $0\%$-$1\%$ steps, we
train policy with RL+GILD, where GILD has not converged; (ii)
RL-only stage: during $1\%$-$100\%$ steps, we train policy with
vanilla RL, where GILD has converged.

\begin{figure}[t]
    \centering
    \includegraphics[width=0.45\linewidth]{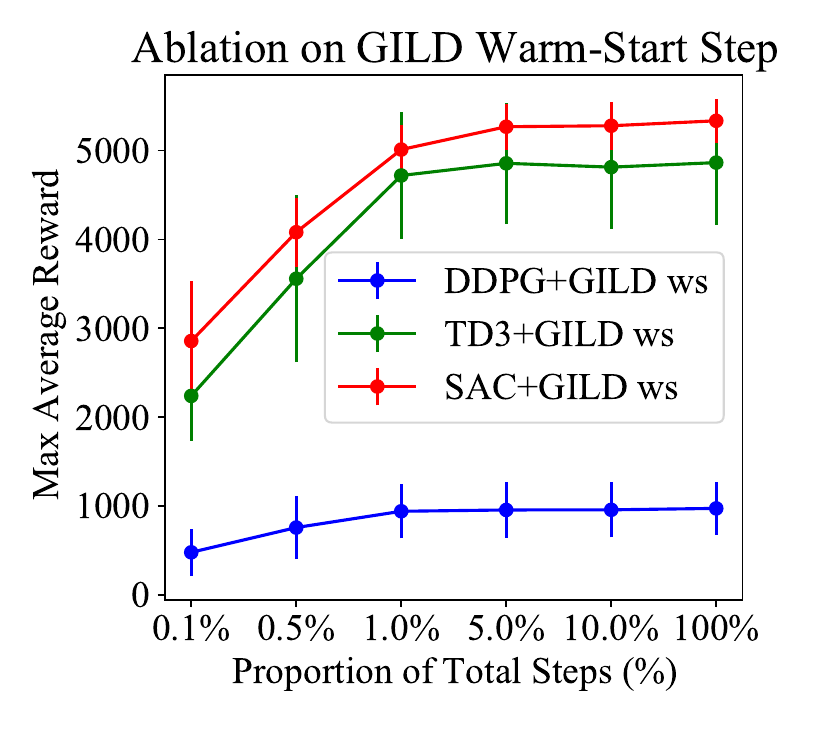}
    \caption{Ablation on warm-start steps. GILD converges within $1\%$ of total steps.}
    \label{fig:ablation on warmstart}
\end{figure}

For example, DDPG+GILD+$1\%$ws trains the policy with RL+GILD at
$0\%$-$1\%$ of total steps and with vanilla DDPG at $1\%$-$100\%$ of
total steps. Figure~\ref{fig:ablation on warmstart} shows the max
average return for three RL+GILD methods trained with different
warm-start steps in the sparse Ant-v2 benchmark. Overall, GILD
converges within $1\%$ of total steps and improves slightly with
more steps, indicating GILD's great potential to enhance RL with
minimal computational cost.

\subsection{RQ4: Computational Efficiency Analysis}
\label{sec: experiments-computational efficiency}

We evaluate the average run time of training each algorithm to
convergence over four MuJoCo tasks, using either author-provided or
open-source implementations. The results are reported in
Table~\ref{tab:run time}. Unsurprisingly, on-policy algorithms take
a longer time to converge due to lower sample efficiency than
off-policy algorithms, especially for LOGO which calculates
KL-divergence at each time step. Although RL+GILD takes longer
training time than RL+IL, RL+GILD with $1\%$ (of total) warm-start
steps significantly reduces training time, while achieving superior
performance (as shown in Figure~\ref{fig:ablation on warmstart}).
Overall, vanilla RL algorithms enhanced with GILD warm-start take
less than half of the computational cost of these (state-of-the-art)
off-policy and on-policy algorithms. We recommend $1\%$ (of total
training steps) as warm-start steps for a minimal increase in
computational cost while significantly improving performance. In
more complex tasks, GILD might converge slower due to a larger
amount of offline data and a higher dimensionality of data (e.g.,
image data).

\begin{table}[t] \footnotesize
  \centering \resizebox{0.9\linewidth}{!}{
  \begin{tabular}{l|ccc|cc}
    \toprule
    \multicolumn{1}{c|}{\multirow{2}{*}{Algorithm}} & \multicolumn{3}{c|}{Off-policy} & \multicolumn{2}{c}{On-policy} \\
    \multicolumn{1}{c|}{}    & DDPG     & TD3      & SAC      & PPO           & LOGO  \\
    \midrule
    Vanilla RL    & \textbf{1h58m}    & \textbf{2h13m}    & \textbf{4h40m}    & 23h51m    & 67h36m \\
    RL+IL & 2h58m     & 3h6m    & 6h4m    & 26h31m   & - \\
    RL+MC & 7h23m    & 7h58m    & 15h47m   & -        & - \\
    RL+GILD & 4h21m    & 4h35m    & 9h24m    & -      & - \\
    RL+GILD+1\%ws & \underline{2h5m}    & \underline{2h18m}    & \underline{4h59m}    & -    & - \\
    \bottomrule
  \end{tabular}
  }
  \caption{Average run time comparison for all methods over four MuJoCo tasks,
``$1\%$ws'' denotes $1\%$ (of total training steps) as warm-start
steps, and ``-'' denotes no such a combination. Off-policy methods
with the shortest time are in bold, and the second shortest are
underlined.}
  \label{tab:run time}
\end{table}

\section{Conclusion}
\label{sec:conclusion}

We develop GILD, a flexible module that meta-learns a general
imitation learning objective function from offline data to enhance
diverse vanilla off-policy RL algorithms with sparse rewards.
Introducing no domain-specific hyperparameter and minimal increase
in computational cost, GILD is intended for diverse vanilla
off-policy RL algorithms. We show that RL+GILD significantly improve
upon baselines in four challenging environments.

\textbf{Limitation and future work.} GILD is conceived within the
single-task meta-RL framework, which necessitates RL agents to learn
from scratch upon encountering unseen tasks. This inherently limits
the extensibility of GILD to few-shot learning scenarios. In future
work, we plan to evolve GILD into the multi-task meta-RL framework,
thereby addressing challenges in few-shot learning paradigms.

\section{Acknowledgements}

This work was supported by the National Natural Science Foundation
of China (No. 62276047 and No. 62176154) and National Foreign Expert
Project of China (No. H20240938).

\bibliography{aaai25}

\appendix

\begin{center}
{\Large\textbf{Supplementary Material}}
\end{center}

\begin{algorithm*}[!h] \small
\caption{DDPG+GILD algorithm} \label{alg:DDPG+GILD} \textbf{Input}:
Actor $\phi$, critic $\theta$, GILD $\omega$, demonstration data
$\mathcal{D}^{\mathrm{dem}}$, and empty replay buffer $\mathcal{D}$
\begin{algorithmic}[1]
    \STATE Initialize target networks with $\theta' \leftarrow \theta$, $\phi' \leftarrow \phi$;

    \FOR{$t=0, 1, \dots, T$}
        \STATE Observe state $s$ and select action $a=\phi(s)+\mathcal{N}$;
        \STATE Execute $a$ in the environment, receive reward $r$ and next state $s'$;
        \STATE Store transition $(s,a,r,s')$ in replay buffer $\mathcal{D}$;
        \STATE Sample a mini-batch of $N$ transitions $(s,a,r,s')$ from $\mathcal{D}$, and $N$ $(s^{\mathrm{d}}, a^{\mathrm{d}})$ from $\mathcal{D}^{\mathrm{dem}}$;

        \STATE \textbf{meta-training:}

        \STATE Update critic by minimizing MSBE loss:
        \begin{equation}
            \theta \leftarrow \theta - \alpha \nabla_{\theta} \frac{1}{N}\sum \Big[Q_{\theta}(s,a) - \Big(r + \gamma Q_{\theta'} \big(s',\phi'(s')\big) \Big) \Big]^{2};
            \nonumber
        \end{equation}

        \STATE Make a copy of actor for pseudo-updating: $\hat{\phi} = \phi$;

        \STATE Calculate RL loss:
        \begin{equation}
            \mathcal{L}^{\mathrm{RL}} (\phi)=- \frac{1}{N} \sum Q_{\theta} \big( s,\phi(s) \big), \quad
            \mathcal{L}^{\mathrm{RL}} (\hat{\phi})=- \frac{1}{N} \sum Q_{\theta} \big( s,\hat{\phi}(s) \big);
            \nonumber
        \end{equation}

        \STATE Calculate conventional imitation learning loss:
        \begin{equation}
            \mathcal{L}^{\mathrm{IL}} (\hat{\phi})=\frac{1}{N} \sum \big(\hat{\phi}(s^{\mathrm{d}})-a^{\mathrm{d}} \big)^{2};
            \nonumber
        \end{equation}

        \STATE Pseudo-update actor with RL+IL:
        \begin{equation}
            \hat{\phi} \leftarrow \hat{\phi} - \alpha \nabla_{\hat{\phi}} \big[\mathrm{w_{rl}} \mathcal{L}^{\mathrm{RL}} (\hat{\phi}) + \mathrm{w_{il}}  \mathcal{L}^{\mathrm{IL}}(\hat{\phi}) \big];
            \nonumber
        \end{equation}

        \STATE Calculate general imitation loss learned by GILD:
        \begin{equation}
            \mathcal{L}^{\mathrm{GILD}}_{\omega}(\phi)=\frac{1}{N} \sum f_{\omega} \big(s^{\mathrm{d}}, a^{\mathrm{d}}, \phi(s^{\mathrm{d}}) \big);
            \nonumber
        \end{equation}

        \STATE Update actor with RL+GILD:
        \begin{equation}
            \phi \leftarrow \phi - \alpha \nabla_{\phi} \big[\mathcal{L}^{\mathrm{RL}} (\phi) + \mathcal{L}^{GILD}_{\omega}(\phi) \big];
            \nonumber
        \end{equation}

        \STATE \textbf{meta-optimization:}
        \STATE Sample a mini-batch of N $s^{\mathrm{val}}$ from $\mathcal{D}$;
        \STATE Calculate meta-loss:
        \begin{equation}
            \mathcal{L}^{\mathrm{meta}}_{\theta} (\phi)= \frac{1}{N} \sum \Big[ \tanh \Big(
            Q_{\theta}\big( s^{\mathrm{val}}, \phi (s^{\mathrm{val}}) \big) - Q_{\theta} \big( s^{\mathrm{val}}, \hat{\phi}(s^{\mathrm{val}}) \big) \Big) \Big] ;
            \nonumber
        \end{equation}

        \STATE Update GILD with meta-loss:
        \begin{equation}
            \omega \leftarrow \omega + \alpha^{2} \frac{\partial \mathcal{L}^{\mathrm{meta}}_{\theta}({\phi}) } {\partial \phi} \cdot
             \frac{\partial^{2} \mathcal{L}^{\mathrm{GILD}}_{\omega}(\phi)}{\partial \phi \partial\omega} ;
        \nonumber
        \end{equation}

        \STATE Update the target networks:
        \begin{gather}
            \theta' \leftarrow \tau \theta + (1-\tau) \theta' \nonumber, \\
            \phi' \leftarrow \tau \phi + (1-\tau) \phi';
            \nonumber
        \end{gather}

    \ENDFOR
\end{algorithmic}
\end{algorithm*}

\begin{algorithm*}[!h] \small
\caption{TD3+GILD algorithm} \label{alg:TD3+GILD} \textbf{Input}:
Actor $\phi$, critic $\theta_{1}, \theta_{2}$, GILD $\omega$,
demonstration data $\mathcal{D}^{\mathrm{dem}}$, and empty replay
buffer $\mathcal{D}$
\begin{algorithmic}[1]

    \STATE Initialize target networks with $\theta_{1}' \leftarrow \theta_{1}$, $\theta_{2}' \leftarrow \theta_{2}$, $\phi' \leftarrow \phi$;

    \FOR{$t=0, 1, \dots, T$}
        \STATE Observe state $s$ and select action $a=\phi(s)+\mathcal{N}$;
        \STATE Execute $a$ in the environment, receive reward $r$ and next state $s'$;
        \STATE Store transition $(s,a,r,s')$ in replay buffer $\mathcal{D}$;
        \STATE Sample a mini-batch of $N$ transitions $(s,a,r,s')$ from $\mathcal{D}$, and $N$ $(s^{\mathrm{d}}, a^{\mathrm{d}})$ from $\mathcal{D}^{\mathrm{dem}}$;

        \STATE Update critic by minimizing MSBE loss:
        \begin{gather}
            \theta_{i} \leftarrow \theta_{i} - \alpha \nabla_{\theta_{i}} \frac{1}{N}\sum \Big[Q_{\theta_{i}}(s,a) - \Big(r + \gamma \min_{j=1,2} Q_{\theta_{j}'} \big(s',\widetilde{a} \big) \Big) \Big]^{2}, \quad \mathrm{for}\ i=1,2, \nonumber \\
            \widetilde{a} = \phi'(s')+ \epsilon,\quad \epsilon \sim clip(\mathcal{N}(0,\sigma), -c, c); \nonumber
        \end{gather}

        \IF{$t$ mod $d$ = 0}

        \STATE \textbf{meta-training:}

        \STATE Make a copy of actor for pseudo-updating: $\hat{\phi} = \phi$;

        \STATE Calculate RL loss:
        \begin{equation}
            \mathcal{L}^{\mathrm{RL}} (\phi)=- \frac{1}{N} \sum Q_{\theta_{1}} \big( s,\phi(s) \big), \quad
            \mathcal{L}^{\mathrm{RL}} (\hat{\phi})=- \frac{1}{N} \sum Q_{\theta_{1}} \big( s,\hat{\phi}(s) \big);
            \nonumber
        \end{equation}

        \STATE Calculate conventional imitation learning loss:
        \begin{equation}
            \mathcal{L}^{\mathrm{IL}} (\hat{\phi})=\frac{1}{N} \sum \big(\hat{\phi}(s^{\mathrm{d}})-a^{\mathrm{d}} \big)^{2};
            \nonumber
        \end{equation}

        \STATE Pseudo-update actor with RL+IL:
        \begin{equation}
            \hat{\phi} \leftarrow \hat{\phi} - \alpha \nabla_{\hat{\phi}} \big[\mathrm{w_{rl}} \mathcal{L}^{\mathrm{RL}} (\hat{\phi}) + \mathrm{w_{il}}  \mathcal{L}^{\mathrm{IL}}(\hat{\phi}) \big];
            \nonumber
        \end{equation}

        \STATE Calculate general imitation loss learned by GILD:
        \begin{equation}
            \mathcal{L}^{\mathrm{GILD}}_{\omega}(\phi)=\frac{1}{N} \sum f_{\omega} \big(s^{\mathrm{d}}, a^{\mathrm{d}}, \phi(s^{\mathrm{d}}) \big);
            \nonumber
        \end{equation}

        \STATE Update actor with RL+GILD:
        \begin{equation}
            \phi \leftarrow \phi - \alpha \nabla_{\phi} \big[\mathcal{L}^{\mathrm{RL}} (\phi) + \mathcal{L}^{GILD}_{\omega}(\phi) \big];
            \nonumber
        \end{equation}

        \STATE \textbf{meta-optimization:}
        \STATE Sample a mini-batch of $N$ $s^{\mathrm{val}}$ from $\mathcal{D}$;
        \STATE Calculate meta-loss:
        \begin{equation}
            \mathcal{L}^{\mathrm{meta}}_{\theta_{1}} (\phi)= \frac{1}{N} \sum \Big[ \tanh \Big(
            Q_{\theta_{1}}\big( s^{\mathrm{val}}, \phi (s^{\mathrm{val}}) \big) - Q_{\theta_{1}} \big( s^{\mathrm{val}}, \hat{\phi}(s^{\mathrm{val}}) \big) \Big) \Big];
            \nonumber
        \end{equation}

        \STATE Update GILD with meta-loss:
        \begin{equation}
            \omega \leftarrow \omega + \alpha^{2} \frac{\partial \mathcal{L}^{\mathrm{meta}}_{\theta_{1}}({\phi}) } {\partial \phi} \cdot
             \frac{\partial^{2} \mathcal{L}^{\mathrm{GILD}}_{\omega}(\phi)}{\partial \phi \partial\omega};
        \nonumber
        \end{equation}

        \STATE Update the target networks:
        \begin{gather}
            \theta_{i}' \leftarrow \tau \theta_{i} + (1-\tau) \theta_{i}', \nonumber \quad \mathrm{for}\ i=1,2, \\
            \phi' \leftarrow \tau \phi + (1-\tau) \phi';
            \nonumber
        \end{gather}

        \ENDIF
    \ENDFOR
\end{algorithmic}
\end{algorithm*}

\begin{algorithm*}[!h] \small
\caption{SAC+GILD algorithm} \label{alg:SAC+GILD} \textbf{Input:}
Actor $\phi$, critic $\theta_{1}, \theta_{2}$, GILD $\omega$,
demonstration data $\mathcal{D}^{\mathrm{dem}}$, empt replay buffer
$\mathcal{D}$, learning rate $\eta$, and temperature $\alpha$
\begin{algorithmic}[1]
    \STATE Initialize target networks with $\theta_{1}' \leftarrow \theta_{1}$, $\theta_{2}' \leftarrow \theta_{2}$;

    \FOR{$t=0, 1, \dots, T$}
        \STATE Observe state $s$ and select action $a \sim \pi_{\phi}(a|s)$;
        \STATE Execute $a$ in the environment, receive reward $r$ and next state $s'$;
        \STATE Store transition $(s,a,r,s')$ in replay buffer $\mathcal{D}$;
        \STATE Sample a mini-batch of $N$ transitions $(s,a,r,s')$ from $\mathcal{D}$, and $N$ $(s^{\mathrm{d}}, a^{\mathrm{d}})$ from $\mathcal{D}^{\mathrm{dem}}$;

        \STATE \textbf{meta-training:}

        \STATE Update critic by minimizing MSBE loss:
        \begin{gather}
            \theta_{i} \leftarrow \theta_{i} - \eta \nabla_{\theta_{i}} \frac{1}{N}\sum \Big[Q_{\theta_{i}}(s,\widetilde{a}) - \Big(r + \big[\gamma \min_{j=1,2} Q_{\theta_{j}'} (s',\widetilde{a} ) - \alpha \log\big(\pi_{\phi}(\widetilde{a}|s')\big) \big] \Big) \Big]^{2}, \  \mathrm{for}\ i=1,2, \nonumber \\
            \widetilde{a} \sim \pi_{\phi}(\widetilde{a}|s'); \nonumber
        \end{gather}

        \STATE Make a copy of actor for pseudo-updating: $\hat{\phi} = \phi$;

        \STATE Calculate RL loss:
        \begin{gather}
            \mathcal{L}^{\mathrm{RL}} (\phi)=\frac{1}{N} \sum \big[ \alpha \log\big(\pi_{\phi}(\widetilde{a}|s) \big) - \min_{i=1,2} Q_{\theta_{i}} \big( s,\widetilde{a}\big) \big], \quad
            \widetilde{a} \sim \pi_{\phi}(\widetilde{a}|s'),
            \nonumber \\
            \mathcal{L}^{\mathrm{RL}} (\hat{\phi})=\frac{1}{N} \sum \big[ \alpha \log\big(\pi_{\hat{\phi}}(\widetilde{a}|s) \big) - \min_{i=1,2} Q_{\theta_{i}} \big( s,\widetilde{a}\big) \big], \quad
            \widetilde{a} \sim \pi_{\hat{\phi}}(\widetilde{a}|s');
            \nonumber
        \end{gather}

        \STATE Calculate conventional imitation learning loss:
        \begin{equation}
            \mathcal{L}^{\mathrm{IL}} (\hat{\phi})=- \frac{1}{N} \sum \log(\pi_{\hat{\phi}} \big( a^{\mathrm{d}}|s^{\mathrm{d}}) \big);
            \nonumber
        \end{equation}

        \STATE Pseudo-update actor with RL+IL:
        \begin{equation}
            \hat{\phi} \leftarrow \hat{\phi} - \eta \nabla_{\hat{\phi}}  \big[\mathrm{w_{rl}} \mathcal{L}^{\mathrm{RL}} (\hat{\phi}) + \mathrm{w_{il}}  \mathcal{L}^{\mathrm{IL}}(\hat{\phi}) \big];
            \nonumber
        \end{equation}

        \STATE Calculate general imitation loss learned by GILD:
        \begin{equation}
            \mathcal{L}^{\mathrm{GILD}}_{\omega}(\phi)=\frac{1}{N} \sum f_{\omega} \big(s^{\mathrm{d}}, a^{\mathrm{d}}, \widetilde{a} \big), \quad \widetilde{a} \sim \pi_{\hat{phi}} (\widetilde{a}|s^{\mathrm{d}}) ;
            \nonumber
        \end{equation}

        \STATE Update actor with RL+GILD:
        \begin{equation}
            \phi \leftarrow \phi - \eta \nabla_{\phi} \big[\mathcal{L}^{\mathrm{RL}} (\phi) + \mathcal{L}^{GILD}_{\omega}(\phi) \big];
            \nonumber
        \end{equation}

        \STATE \textbf{meta-optimization:}
        \STATE Sample a mini-batch of $N$ $s^{\mathrm{val}}$ from $\mathcal{D}$;
        \STATE Calculate meta-loss:
        \begin{equation}
            \mathcal{L}^{\mathrm{meta}}_{\theta} (\phi)= \frac{1}{N} \sum \Big[ \tanh \Big(
            \min_{i=1,2} Q_{\theta_{i}} \big( s^{\mathrm{val}}, \phi (s^{\mathrm{val}}) \big) - \min_{i=1,2} Q_{\theta_{i}} \big( s^{\mathrm{val}}, \hat{\phi}(s^{\mathrm{val}}) \big) \Big) \Big] ;
            \nonumber
        \end{equation}

        \STATE Update GILD with meta-loss:
        \begin{equation}
            \omega \leftarrow \omega + \eta^{2} \frac{\partial \mathcal{L}^{\mathrm{meta}}_{\theta}({\phi}) } {\partial \phi} \cdot
             \frac{\partial^{2} \mathcal{L}^{\mathrm{GILD}}_{\omega}(\phi)}{\partial \phi \partial\omega} ;
        \nonumber
        \end{equation}

        \IF{$t$ mod $d$ = 0}

        \STATE Update the target networks:
        \begin{equation}
            \theta_{i}' \leftarrow \tau \theta_{i} + (1-\tau) \theta_{i}', \nonumber \quad \mathrm{for}\ i=1,2; \nonumber
        \end{equation}

        \ENDIF
    \ENDFOR
\end{algorithmic}
\end{algorithm*}

\section{Practical Algorithms}
\label{app:algorithm}

We provide practical algorithms for three vanilla off-policy RL
algorithms enhanced with GILD, which are DDPG+GILD in
Algorithm~\ref{alg:DDPG+GILD}, TD3+GILD in
Algorithm~\ref{alg:TD3+GILD} and SAC+GILD in
Algorithm~\ref{alg:SAC+GILD}.

\section{Implementation Details}
\label{app:implementation details}

The experiments are run on a computer with Intel Xeon Silver 4214R
CPU with max CPU speed of 2.40GHz. We implement all the algorithms
in this paper using PyTorch. All algorithms are run with a single
Nvidia GeForce RTX 3090 GPU.

To ensure a fair and identical experimental evaluation across
algorithms, we train RL+IL and RL+GILD algorithms using
hyperparameters same as the vanilla RL algorithms. We train each
off-policy algorithm for 1 million steps in sparse reward
environment and evaluate it every 5000 steps with dense rewards.
On-policy algorithms are trained with much more steps (e.g., 30
million) to ensure convergence. Results are averaged over five
random seeds and standard deviation of evaluation reward is shown
with shaded region or error bar.

\textbf{Hyperparameters and network structure.} We implement GILD as
a three-layer (256$\times$256) fully connected network with ReLu
activation functions in hidden layers and SoftPlus activation
functions in the output layer. We report the complete
hyperparameters for DDPG family (DDPG, DDPG+IL, DDPG+GILD), TD3
family (TD3, TD3+IL, TD3+GILD), and SAC family (SAC, SAC+IL,
SAC+GILD) algorithms in Table~\ref{tab:hyperparameter}. All
algorithms that are in the same family share the same
hyperparameters and we include no domain-specific parameters.

\begin{table*}[t] \footnotesize
  \centering
  \begin{tabular}{llll}
    \toprule
    Parameter     & DDPG    & TD3   & SAC \\
    \midrule
        Optimizer      & Adam   & Adam  & Adam \\
        Learning rate  & $3 \cdot 10^{-4}$  & $3 \cdot 10^{-4}$ & $3 \cdot 10^{-4}$ \\
        Discount ($\gamma$)  & 0.99 & 0.99  & 0.99 \\
        Replay buffer size  & $2 \cdot 10^{6}$  & $2 \cdot 10^{6}$  & $2 \cdot 10^{6}$ \\
        Number of hidden layers  & 2    & 2     & 2 \\
        Number of hidden units per layer  & 256    & 256   & 256\\
        Activation function (hidden layer)  & ReLU     & ReLU      & ReLU \\
        Activation function (actor output layer) & Tanh    & Tanh  & Tanh \\
        Target update rate ($\tau$)  & $5 \cdot 10^{-3} $   & $5 \cdot 10^{-3} $    & $5 \cdot 10^{-3} $ \\
        Batch size  & 256  & 256   & 256 \\
        Exploration noise   & $\mathcal{N}(0, 0.2)$ & $\mathcal{N}(0, 0.2)$ & - \\
        Policy noise     & -    & 0.2   & - \\
        Noise clip     & -    & 0.5   & - \\
        target update interval   & -    & 2     & 2 \\
        Temperature ($\alpha$)     & -    & -  & 0.2 \\
    \bottomrule
  \end{tabular}
  \caption{Hyperparameters for the DDPG, TD3, and SAC family algorithms.}
  \label{tab:hyperparameter}
\end{table*}

\textbf{Demonstration data details.} Following LOGO, we train three
vanilla RL algorithms in the dense reward environment to provide
optimal baselines, and use the partially trained expert that is
still at a sub-optimal stage of learning to provide behavior data
for both RL+IL and RL+GILD. We provide details on the demonstration
data collected using the Behavior policy in
Table~\ref{tab:demonstration}.

\begin{table*}[t] \footnotesize
  \centering
  \begin{tabular}{ccccccc}
    \toprule
    \multirow{2}{*}{Algorithm} & \multicolumn{2}{c}{DDPG} & \multicolumn{2}{c}{TD3}  & \multicolumn{2}{c}{SAC}  \\
    & Samples & Average Return & Samples & Average Return & Samples & Average Return \\
    \midrule
    Hopper-v2                  & 3719    & 1085.80        & 3489    & 1198.65        & 10000   & 1858.09        \\
    Walker-v2                  & 3612    & 1130.45        & 4627    & 1453.35        & 5156    & 1883.53        \\
    HalfCheetah-v2             & 10000   & 4049.00        & 10000   & 3731.08        & 10000   & 4088.43        \\
    Ant-v2                     & 5511    & 243.80         & 10000   & 1807.00        & 8628    & 1926.96        \\
    \bottomrule
  \end{tabular}
  \caption{Demonstration data details.}
  \label{tab:demonstration}
\end{table*}

\textbf{Benchmark and algorithm licenses.} We adopt four MuJoCo
environments (Hopper-v2, Walker2d-v2, HalfCheetah-v2 and Ant-v2)
from OpenAI, which has an MIT license. All algorithms are run with
their official GitHub repositories. ``OurDDPG'', TD3, SAC, and PPO
have MIT license. Meta-Critic and LOGO have CC-BY 4.0 license.

\section{Full Results in RQ1}
\label{app:full results in RQ1}

This section includes full results for RQ1, including (i) learning
curves of DDPG+GILD, TD3+GILD, SAC+GILD, PPO, PPO+IL, LOGO, and
Meta-Critic; (ii) Max average returns of Expert, Behavior, and
Meta-Critic; (iii) Average return of Expert and Behavior for three
vanilla off-policy RL algorithms (DDPG, TD3, and SAC); (iv) Average
normalized scores of three vanilla RL algorithms, their
corresponding RL+IL and RL+GILD variants, and Behavior baselines.

\begin{table*}[t] \footnotesize
\centering
\begin{tabular}{lccccc}
\hline \multicolumn{2}{c}{Algorithm}    & Hopper  & Walker  &
HalfCheetah & Ant     \\ \hline
\multirow{2}{*}{DDPG} & Expert   & 2500.18 & 2239.44 & 9678.32     & 998.41  \\
                      & Behavior & 1085.80 & 1130.45 & 4049.00     & 243.80  \\ \hline
\multirow{2}{*}{TD3}  & Expert   & 3370.08 & 3870.46 & 10002.19    & 3726.70 \\
                      & Behavior & 1198.65 & 1453.35 & 3731.08     & 1807.00 \\ \hline
\multirow{2}{*}{SAC}  & Expert   & 3491.95 & 4069.22 & 10846.69    & 5000.17 \\
                      & Behavior & 1858.09 & 1883.53 & 4088.43     & 1926.96 \\ \hline
\end{tabular}
\caption{Average return of Expert and Behavior models for three
vanilla off-policy RL algorithms (DDPG, TD3, SAC). All results are
averaged over 10 evaluation episodes in dense reward environment.}
\label{tab:max-avg-return for Expert and Behavior}
\end{table*}

\begin{figure*}[t]
    \centering
    \includegraphics[width=1\linewidth]{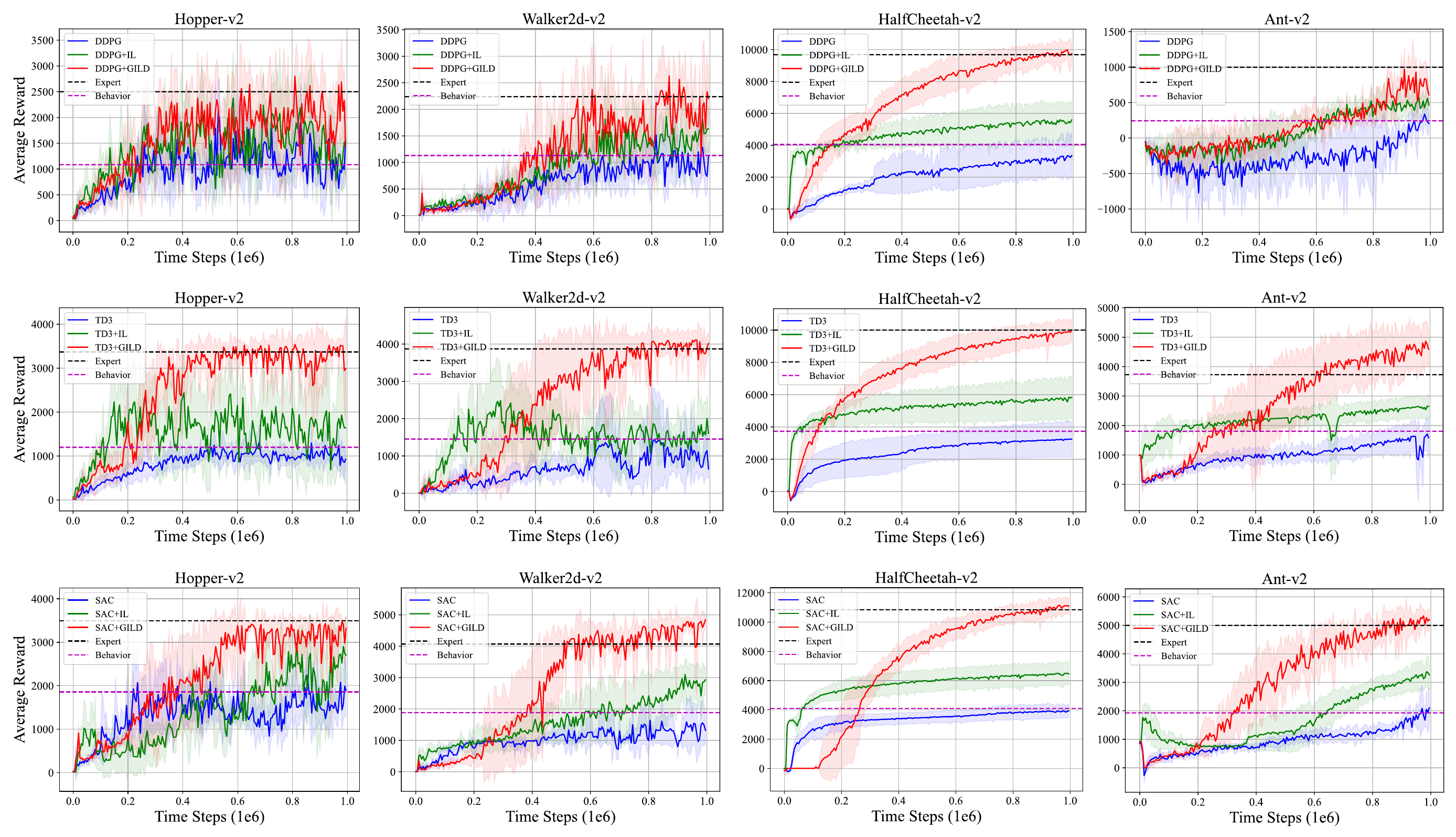}
    \caption{Learning curve with Mean-STD of vanilla RL algorithms (DDPG, TD3, and SAC), and their corresponding RL+IL and RL+GILD variants in four MuJoCo tasks with sparse rewards.}
    \label{fig:learning curve: 3RL+GILD}
\end{figure*}

\begin{figure*}[t]
    \centering
    \includegraphics[width=1\linewidth]{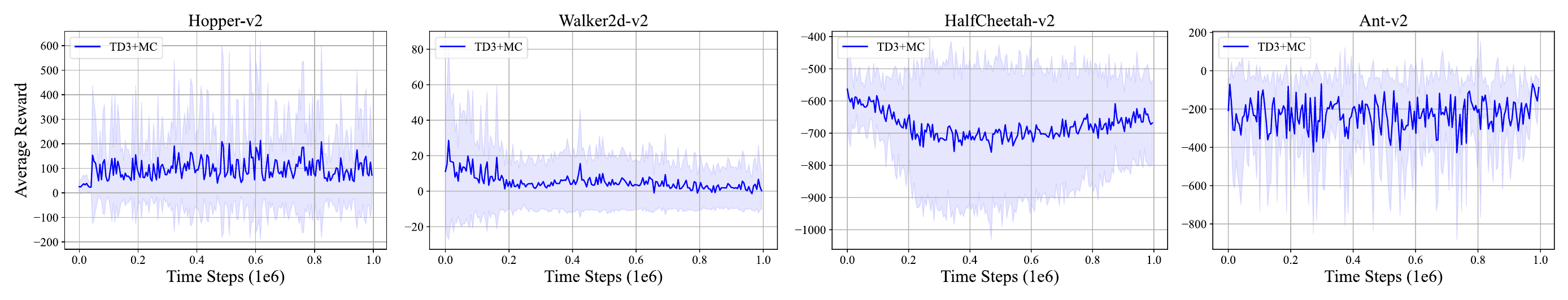}
    \caption{Learning curve with Mean-STD of TD3+MC in four MuJoCo tasks with sparse rewards.}
    \label{fig:learning curve: TD3+MC}
\end{figure*}

\begin{figure*}[t]
    \centering
    \includegraphics[width=1\linewidth]{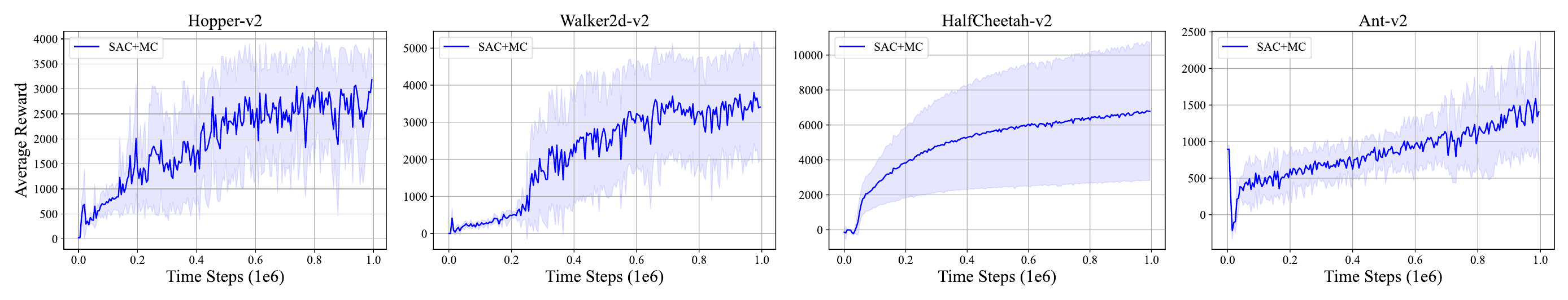}
    \caption{Learning curve with Mean-STD of SAC+MC in four MuJoCo tasks with sparse rewards.}
    \label{fig:learning curve: SAC+MC}
\end{figure*}

The average return of Expert and Behavior for three vanilla
off-policy RL algorithms (DDPG, TD3, SAC) are reported in
Table~\ref{tab:max-avg-return for Expert and Behavior}. We display
the learning curves of vanilla RL algorithms (DDPG, TD3 and SAC)
with their corresponding RL+IL and RL+GILD variants in
Figure~\ref{fig:learning curve: 3RL+GILD}. Learning curves of
DDPG+MC, TD3+MC and SAC+MC are shown in Figure~\ref{fig:learning
curve: DDPG+MC}, Figure~\ref{fig:learning curve: TD3+MC} and
Figure~\ref{fig:learning curve: SAC+MC}, respectively.

\begin{figure*}[t]
    \centering
    \includegraphics[width=1\linewidth]{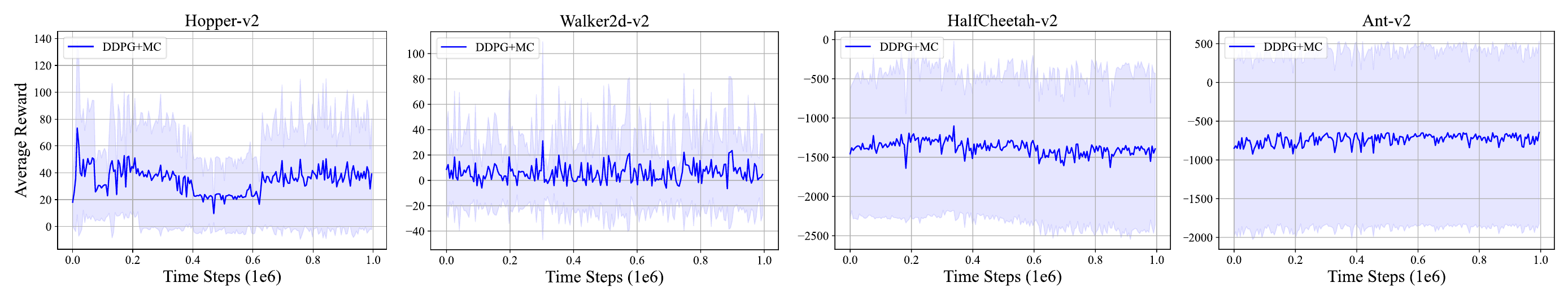}
    \caption{Learning curve with Mean-STD of DDPG+MC in four MuJoCo tasks with sparse rewards.}
    \label{fig:learning curve: DDPG+MC}
\end{figure*}

Meta-Critic includes DDPG+MC, TD3+MC, and SAC+MC. We choose the one
(i.e., SAC+MC) with the highest max average return to be reported in
the main paper. To ensure a fair and identical experimental
evaluation, we re-run DDPG-MC-sa, TD3-MC-sa, SAC-MC-sa, which are
augmented with state-action feature, using the author-provided
implementation\footnote{https://github.com/zwfightzw/Meta-Critic}.
Max average return of Meta-Critic is reported in
Table~\ref{tab:max-avg-return for Meta-Critic}.

\begin{table*}[t] \footnotesize
  \centering
  \begin{tabular}{lcccc}
    \toprule
    Algorithm   & Hopper-v2    & Walker2d-v2 & HalfCheetah-v2   & Ant-v2 \\
    \midrule
    DDPG+MC        & 73.3+-64.6  & 31.2+-78.1  & -1099.7+-1079.0  & -644.8+-1177.0 \\
    TD3+MC     & 214.0+-403.7  & 28.6+-55.8  & -562.9+-108.5  & -67.7+-9.9 \\
    SAC+MC    & \textbf{3185.2+-526.9}  & \textbf{3807.0+-1377.1}  & \textbf{6811.6+-3981.0}  & \textbf{1588.9+-782.8} \\
    \midrule
    Meta-Critic     & 3185.2+-526.9  & 3807.0+-1377.1  & 6811.6+-3981.0  & 1588.9+-782.8 \\
    \bottomrule
  \end{tabular}
  \caption{Max average return of Meta-Critic, which includes DDPG+MC, TD3+MC,
SAC+MC. Results are run on sparse environments over 5 trials,
``$\pm$'' captures the standard deviation over trials. We choose the
one (i.e., SAC+MC) with maximum max average return to be reported in
the main paper.}
  \label{tab:max-avg-return for Meta-Critic}
\end{table*}

\begin{figure*}[t]
    \centering
    \includegraphics[width=1\linewidth]{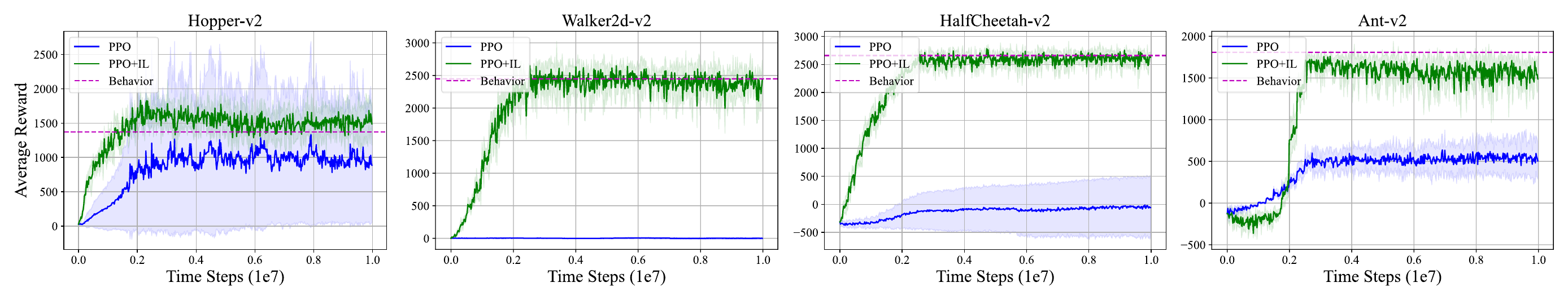}
    \caption{Learning curve with Mean-STD of PPO and PPO+IL in four MuJoCo tasks with sparse rewards.}
    \label{fig:learning curve: PPO}
\end{figure*}

We use an open-source implementation of
PPO\footnote{https://github.com/nikhilbarhate99/PPO-PyTorch}.
Considering PPO is an on-policy alike LOGO, we use the demonstration
data in LOGO for sparse Hopper-v2, Walker2d-v2 and HalfCheetah-v2.
For sparse Ant-v2 benchmark that is not evaluated in LOGO, we
provide demonstration data collected by the Behavior policy trained
with SAC in dense environment. This demonstration data is also used
by SAC+IL and SAC+GILD. Learning curves of PPO and PPO+IL are shown
in Figure~\ref{fig:learning curve: PPO}.

\begin{figure*}[ht]
    \centering
    \includegraphics[width=1\linewidth]{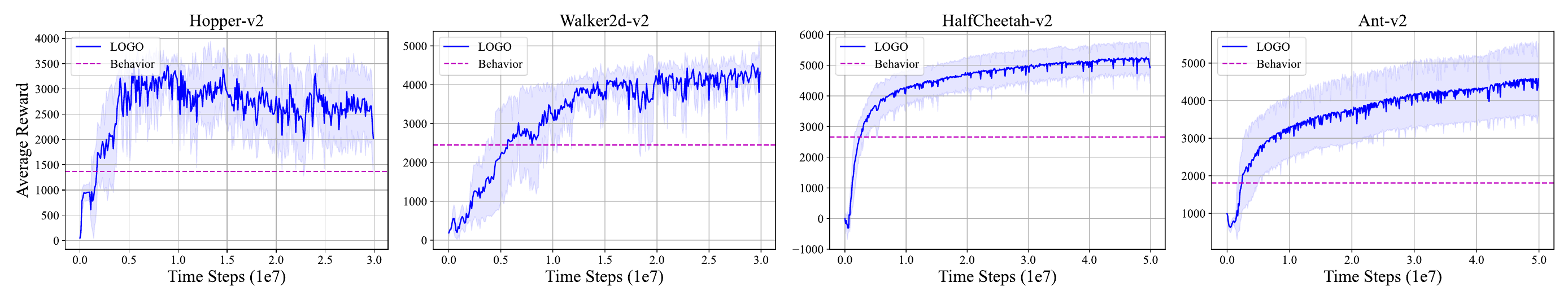}
    \caption{Learning curve with Mean-STD of LOGO in four MuJoCo tasks with sparse rewards.}
    \label{fig:learning curve: LOGO}
\end{figure*}

We re-run LOGO using the author-provided
implementation\footnote{https://github.com/DesikRengarajan/LOGO}.
For sparse Ant-v2 benchmark that is not evaluated in LOGO, we
provide demonstration data collected by the Behavior policy trained
with SAC in dense environment. This demonstration data is also used
by SAC+IL and SAC+GILD. Learning curves of LOGO is shown in
Figure~\ref{fig:learning curve: LOGO}..

We report the complete average normalized scores of three vanilla
off-policy RL algorithms (DDPG, TD3, and SAC) and their
corresponding variants (RL+IL and RL+GILD) and Behavior algorithms
in Table~\ref{tab:Avg_Norm_Score summary}.

\begin{table*}[t]
  \centering
  \begin{tabular}{lcccc}
    \toprule
    Algorithm   & Hopper-v2    & Walker2d-v2 & HalfCheetah-v2   & Ant-v2 \\
    \midrule
    DDPG Behavior    & 43.43  & 50.48  & 41.84  & 24.42 \\
    DDPG        & 84.91$\pm$23.63  & 67.85$\pm$39.38  & 34.60$\pm$15.39  &  33.96$\pm$10.94 \\
    DDPG+IL     & 95.13$\pm$36.24  & 83.41$\pm$21.86  & 57.90$\pm$11.67  & 57.60$\pm$21.58 \\
    DDPG+GILD (ours)    & \underline{\textbf{112.15$\pm$9.42}}  & \underline{117.54$\pm$16.66}  & \underline{\textbf{103.20$\pm$5.29}}  & \underline{97.32$\pm$29.72} \\
    \midrule
    TD3 Behavior    & 35.57  & 37.55  & 37.30  & 48.49 \\
    TD3         & 39.19$\pm$12.28  & 36.86$\pm$36.51  & 32.51$\pm$11.35  & 45.95$\pm$15.10 \\
    TD3+IL      & 72.34$\pm$26.42  &  64.30$\pm$23.35  &  58.43$\pm$13.21  & 71.38$\pm$10.61 \\
    TD3+GILD (ours)     & \underline{105.00$\pm$3.10}  & \underline{106.28$\pm$7.25}  & \underline{99.96$\pm$7.55}  & \underline{\textbf{130.53$\pm$18.76}} \\
    \midrule
    SAC Behavior    & 53.21  & 46.29  & 37.69  & 38.54 \\
    SAC         & 64.01$\pm$16.31  & 40.38$\pm$19.89  & 36.38$\pm$4.47  & 42.14$\pm$14.36 \\
    SAC+IL      & 85.61$\pm$7.54  & 76.23$\pm$11.71  & 59.96$\pm$7.40  & 67.42$\pm$9.33 \\
    SAC+GILD (ours)     & \underline{99.39$\pm$2.44}  & \underline{\textbf{118.95$\pm$5.99}}  & \underline{102.90$\pm$5.10}  & \underline{106.70$\pm$4.94} \\
    \bottomrule
  \end{tabular}
    \caption{Average normalized scores of three vanilla off-policy RL algorithms,
their corresponding variants (RL+IL and RL+GILD), and Behavior
algorithms. The scores are normalized using the max average return
of Expert (with a score of 100). Results are run on sparse
environments over 5 trials, and ``$\pm$'' captures the standard
deviation over trials. Max value for each category is underlined,
and max value overall is in bold.}
  \label{tab:Avg_Norm_Score summary}
\end{table*}

\section{Further Visualization Results}
\label{app:further visualization}

We apply Principal Component Analysis (PCA) to reduce the dimension
of policy parameter $\phi$ and extract the top-2 representative
principle components for plotting on the 2D surface. Every point on
the surface represents a policy. These policies are densely
evaluated over 10 episodes to get the average reward values, which
serve as colors of the points.

\begin{figure*}[t]
    \centering
    \includegraphics[width=1.0\linewidth]{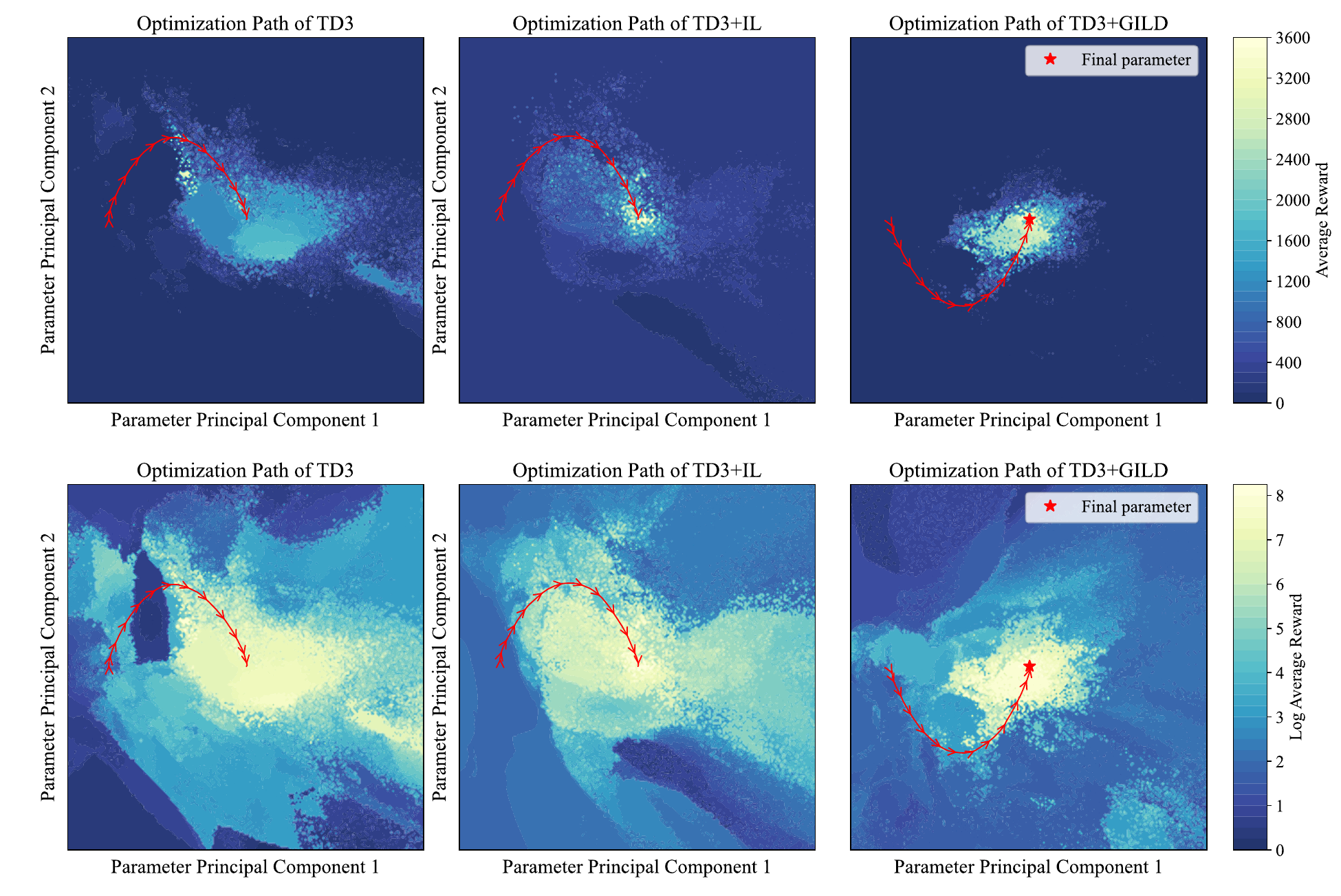}
    \caption{Visualization of optimization path for TD3 (left), TD3+IL (middle),
TD3+GILD (right) in sparse Hopper-v2 environment. The red star
denotes the final parameter point. Both top and bottom figures are
plotted on the same average reward surface, with the only difference
of values (colors) to plot due to log operation.}
    \label{fig:Opt_path for TD3s in Hopper}
\end{figure*}

Figure~\ref{fig:Opt_path for TD3s in Hopper} shows further
visualization result on the policy optimization path of TD3, TD3+IL
and TD3+GILD in sparse Hopper-v2 environment. The red star denotes
the final parameter point. For better understanding, we display both
raw average-return (top) to distinct TD3+GILD from the others, and
log-average-return (bottom) to present the detailed situation of the
average reward surface. Both top and bottom figures are plotted on
the same average reward surface, with the only difference of values
(colors) to plot due to log operation.

The optimization path demonstrates that (i) vanilla TD3 struggles to
find a parameter area with higher reward and finally learns a bad
parameter (policy); (ii) TD3+IL finds a parameter area with higher
reward more quickly via imitating sub-optimal behaviors in
demonstration data, while eventually stuck in the sub-optimal area
due to restriction from supervised IL objective function; (iii)
Although TD3+GILD moves slower than TD3+IL with implicit guidance
from the meta-learned loss function given by GILD, it finally finds
a parameter with the highest reward (i.e., optimal policy) by
distilling knowledge from sub-optimal demonstrations instead of
explicitly imitation.

\section{Further Run Time Results}

Table~\ref{tab:avg run time} shows the run time of training each
algorithm to convergence over four MuJoCo tasks.

\begin{table*}[t] \footnotesize
  \centering
  \begin{tabular}{lccccc}
    \toprule
    Algorithm   & Hopper-v2    & Walker2d-v2 & HalfCheetah-v2   & Ant-v2  & Average \\
    \midrule
    DDPG        & \textbf{1h50m}   & \textbf{1h51m}  & \textbf{2h1m}  & \textbf{2h10m} & \textbf{1h58m} \\
    DDPG+IL     & 2h54m  & 2h52m  & 2h59m  & 3h7m & 2h58m \\
    DDPG+MC     & 7h13m  & 7h17m  & 7h26m  & 7h36m & 7h23m \\
    DDPG+GILD   & 4h12m  & 4h12m  & 4h26m  & 4h34m  & 4h21m \\
    DDPG+GILD+1\%ws    & \underline{1h58m}  & \underline{1h59m}  & \underline{2h8m} & \underline{2h15m}  & \underline{2h5m} \\
    \midrule
    TD3         & \textbf{2h7m}  & \textbf{2h8m}  & \textbf{2h14m}  & \textbf{2h23m}  & \textbf{2h13m} \\
    TD3+IL      & 3h3m  & 3h1m  & 3h5m  & 3h15m  & 3h6m \\
    TD3+MC      & 7h48m  & 7h54m  & 8h0m  & 8h10m  & 7h58m \\
    TD3+GILD    & 4h26m  & 4h27m  & 4h39m  & 4h48m & 4h35m \\
    TD3+GILD+1\%ws    & \underline{2h9m}  & \underline{2h10m}  & \underline{2h21m}  & \underline{2h32m} & \underline{2h18m} \\
    \midrule
    SAC         & \textbf{4h33m}  & \textbf{4h34m}  & \textbf{4h41m}  & \textbf{4h52m}  & \textbf{4h40m} \\
    SAC+IL      & 5h58m  & 5h57m  & 6h3m  & 6h18m  & 6h4m \\
    SAC+MC      & 15h41m  & 15h43m  & 15h45m  & 15h59m  & 15h47m \\
    SAC+GILD    & 9h12m  & 9h13m & 9h30m  & 9h41m & 9h24m  \\
    SAC+GILD+1\%ws    & \underline{4h46m}  & \underline{4h43m} & \underline{5h8m}  & \underline{5h19m} & \underline{4h59m}  \\
    \midrule
    PPO         & \textbf{23h43m}  & \textbf{23h44m}  & \textbf{23h53m}  & \textbf{24h4m}  & \textbf{23h51m} \\
    PPO+IL      & \underline{26h19m}  & \underline{26h23m}  & \underline{26h35m}  & \underline{26h47m}  & \underline{26h31m} \\
    LOGO        & 67h27m  & 67h28m  & 67h38m & 67h51m & 67h36m \\
    \bottomrule
  \end{tabular}
  \caption{Run time comparison for all methods on four tasks. Methods with the
shortest time in their category are in bold, and the second shortest
are underlined.}
  \label{tab:avg run time}
\end{table*}

\newpage

\end{document}